\documentclass[10pt,twocolumn,letterpaper]{article}

\usepackage[pagenumbers]{cvpr} 

\usepackage{graphicx}
\usepackage{amsmath}
\usepackage{amssymb}
\usepackage{booktabs}

\usepackage{graphbox}

\usepackage{makecell}
\usepackage{tabularx}
\usepackage{multirow}
\usepackage{multicol}

\usepackage[T1]{fontenc}    
\usepackage{amsfonts}       
\usepackage{pifont}
\usepackage{dsfont}

\usepackage{algorithm}
\usepackage{algorithmic}

\usepackage{color}
\usepackage[dvipsnames,svgnames,table]{xcolor}

\usepackage[utf8]{inputenc} 
\usepackage{url}            
\usepackage{mathtools}
\usepackage{nicefrac}       
\usepackage[accsupp]{axessibility} 

\usepackage[normalem]{ulem}

\long\def\ignorethis#1{}

\newcommand{\paren}[1]{\left( #1 \right)}

\newcommand{\Paragraph}[1]{\vspace{1mm}\noindent\textbf{#1}}

\newcommand{\ignore}[1]{}   

\newcommand{\figspace}{\vspace{-2mm}}
\newcommand{\figxspace}{\vspace{-3mm}}

\definecolor{BurntOrange}{rgb}{0.8,0.33,0.0}
\definecolor{DarkBlue}{rgb}{0.0,0.0,0.75}

\newcolumntype{L}[1]{>{\raggedright\let\newline\\\arraybackslash\hspace{0pt}}m{#1}}
\newcolumntype{C}[1]{>{\centering\let\newline\\\arraybackslash\hspace{0pt}}m{#1}}
\newcolumntype{R}[1]{>{\raggedleft\let\newline\\\arraybackslash\hspace{0pt}}m{#1}}

\definecolor{cvprblue}{rgb}{0.21,0.49,0.74}
\usepackage[pagebackref,breaklinks,colorlinks,citecolor=cvprblue,hypertexnames=false]{hyperref}

\usepackage[capitalize]{cleveref}
\crefname{section}{Sec.}{Secs.}
\Crefname{section}{Section}{Sections}
\Crefname{table}{Table}{Tables}
\crefname{table}{Tab.}{Tabs.}

\def\paperID{8206} 
\def\confName{CVPR}
\def\confYear{2024}

\title{Beyond Image Super-Resolution for Image Recognition \\ with Task-Driven Perceptual Loss}

\begin{document}

\author{$\text{Jaeha Kim}^{1}$\quad $\text{Junghun Oh}^{1}$\quad $\text{Kyoung Mu Lee}^{1,2}$\\
$^{1}$Dept. of ECE\&ASRI, $^{2}$IPAI, Seoul National University, Korea\\
{\tt\small jhkim97s2@gmail.com, \{dh6dh, kyoungmu\}@snu.ac.kr}
}
\maketitle
\begin{abstract}
In real-world scenarios, image recognition tasks, such as semantic segmentation and object detection, often pose greater challenges due to the lack of information available within low-resolution (LR) content.
Image super-resolution (SR) is one of the promising solutions for addressing the challenges.
However, due to the ill-posed property of SR, it is challenging for typical SR methods to restore task-relevant high-frequency contents, which may dilute the advantage of utilizing the SR method.
Therefore, in this paper, we propose \textbf{S}uper-\textbf{R}esolution \textbf{f}or \textbf{I}mage \textbf{R}ecognition (SR4IR) that effectively guides the generation of SR images beneficial to achieving satisfactory image recognition performance when processing LR images.
The critical component of our SR4IR is the task-driven perceptual (TDP) loss that enables the SR network to acquire task-specific knowledge from a network tailored for a specific task.
Moreover, we propose a cross-quality patch mix and an alternate training framework that significantly enhances the efficacy of the TDP loss by addressing potential problems when employing the TDP loss.
Through extensive experiments, we demonstrate that our SR4IR achieves outstanding task performance by generating SR images useful for a specific image recognition task, including semantic segmentation, object detection, and image classification.
The implementation code is available at \href{https://github.com/JaehaKim97/SR4IR}{https://github.com/JaehaKim97/SR4IR}.
\end{abstract}

\section{Introduction}
\label{sec:introduction}
\begin{figure}[t!]
    \centering
    \subfloat[Ground-truth]{\includegraphics[width=0.48\linewidth]{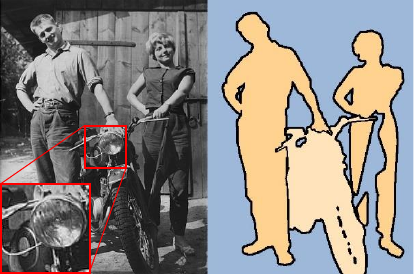}}
    \hfill
    \subfloat[Bilinear-up]{\includegraphics[width=0.48\linewidth]{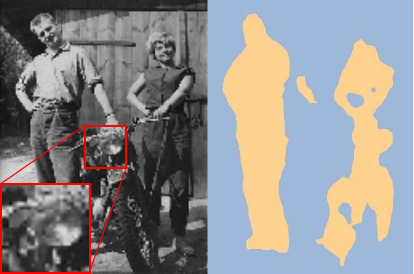}}
    \\
    \vspace{2mm}
    \subfloat[SwinIR~\cite{sr_swinir}]{\includegraphics[width=0.48\linewidth]{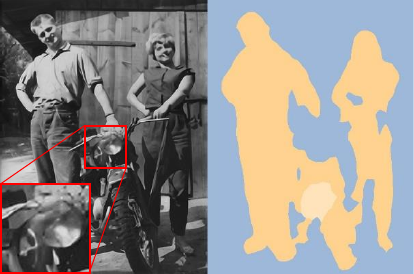}}
    \hfill
    \subfloat[SwinIR\cite{sr_swinir}~+~\textbf{SR4IR(Ours)}]{\includegraphics[width=0.48\linewidth]{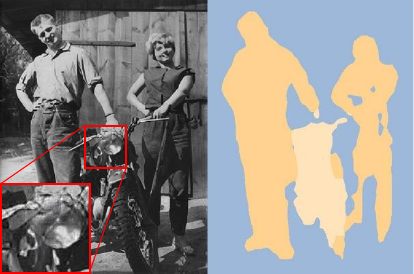}}
    \\
    \figspace
    \caption{
        \textbf{Visualizations and comparisons of the results on semantic segmentation task with PASCAL VOC dataset~\cite{pascal-voc-2012}.}
        (a) The ground-truth image and the class label map.
        (b) The bilinear-upsampled image and the predicted result when trained on bilinear-upsampled images.
        %
        (c) the SR result of the SwinIR~\cite{sr_swinir} and the predicted result when trained on the SR images.
        (d) the SR result of the SwinIR and the predicted result when trained by the proposed SR4IR framework.
        %
        We use DeepLabV3~\cite{chen2017rethinking} as a task network and the downsampling scale factor is x4.
    }
    \label{fig:main_figure}
    \figxspace
    \vspace{-3mm}
\end{figure}

After the pioneering AlexNet~\cite{cls_alexnet}, numerous deep neural network-based methods have achieved superior performance in various image recognition tasks, such as image classification~\cite{cls_resnet, cls_conv2020s, net_densenet, cls_effnet, net_mobilenet, net_vgg, misc_vit}, object detection~\cite{det_fasterrcnn, det_retinanet, det_ssd, det_yolo}, and semantic segmentation~\cite{seg_deeplab, seg_fcn, zhang2022topformer, xie2021segformer}.
%
In real-world applications, such as medical image recognition and surveillance systems, a network for a task, which we refer to as a \textit{task network}, may encounter input images containing low-resolution (LR) contents due to various factors, including limitations of cameras or storage capabilities and the presence of tiny objects.
In this circumstance, achieving satisfactory task performance is limited by the lack of high-frequency components crucial for a task.
%
%

In this paper, we address these challenges by employing single image super-resolution~(SR) methods, as SR can serve as a prominent pre-processing tool to restore missing high-frequency details in LR images.
Conventionally, SR aims to restore high-resolution (HR) images mainly via pixel-wise loss~\cite{sr_edsr, sr_swinir, sr_rdn, sr_vdsr, sr_dbpn, rs_ircnn} or perceptual loss~\cite{misc_perceptual,bruna2016super,leon2015_texture}.
However, due to the ill-posed nature of SR, these approaches may not ensure the reconstruction of features crucial for achieving better performance for a specific task, thereby leading to marginal performance improvement from SR.
Thus, in this paper, our primary goal is to develop a comprehensive framework, named \textbf{S}uper-\textbf{R}esolution \textbf{f}or \textbf{I}mage \textbf{R}ecognition (SR4IR), wherein the SR and task network interplay harmoniously to generate SR images, especially focusing on incorporating task-relevant features, which are exploited to enhance task performance.

To attain our goal, we focus on leveraging the concept of perceptual loss, as it is devised to encourage SR images to mimic the HR counterparts within a feature space.
Typically, the perceptual loss is computed using VGGNet~\cite{net_vgg} pre-trained on ImageNet dataset~\cite{data_imagenet}.
However, we note that employing perceptual loss in the conventional manner results in modest improvements in task performance.
Instead, we propose a Task-Driven Perceptual (TDP) loss that enables the emulation of HR images within the feature space of a task network.
By doing so, our TDP loss facilitates the restoration of high-frequency details, especially beneficial for enhancing task performance.


%
%
%

The effectiveness of our TDP loss is significantly influenced by the knowledge that the task network acquires.
%
%
For high-level vision tasks, it is widely acknowledged that a network can fall into learning non-intrinsic features, leading to biased feature representations~\cite{geirhos2020shortcut}.
A network possessing biased knowledge on a task can diminish the efficacy of the TDP loss, as it may fail to deliver meaningful signals to the SR network for the restoration of other valuable features.
%
%
%
To tackle the problem, we introduce a novel data augmentation strategy for training the task network, called Cross-Quality patch Mix~(CQMix), which blends SR and HR images at the patch level to randomly eliminate high-frequency components.
Consequently, CQMix can prevent a task network from learning shortcut features, thereby allowing us to fully harness the benefits of TDP loss.
%

%

%


In addition to calculating perceptual loss using a task network, we introduce further variations to the perceptual loss by applying it to a task network undergoing training instead of a pre-trained one.
This strategy substantially improves the effectiveness of our TDP loss, since a pre-trained (then fixed) task network cannot learn to produce meaningful features from SR images.
To this end, we constitute the SR4IR as an alternate training framework that trains the SR and task network alternately throughout the training procedure.
In the first phase, we train the SR network using the TDP loss with the temporarily frozen task network.
Subsequently, we train only the task network with task-specific loss using data augmented by the CQMix.
To evaluate the proposed methods, we have examined the integration of SR with three representative image recognition tasks: semantic segmentation, object detection, and image classification.
For each task, we evaluate our method on two representative SR architectures, EDSR~\cite{sr_edsr} and SwinIR~\cite{sr_swinir}, with scale factors of $\times$4 and $\times$8.
%
Our extensive experiments clearly demonstrate that our SR4IR significantly improves task performance while achieving perceptually appealing SR results compared to baseline methods, as shown in Figure~\ref{fig:main_figure}.
These results indicate that our framework can be generally adopted in various applications.
%
%
We summarize our contributions as follows:
\vspace{1mm}
\begin{itemize}
    %
    \item To our knowledge, we are the first to introduce a comprehensive SR framework that addresses challenges posed by LR contents across various image recognition tasks.
    \vspace{2mm}
    \item We propose the task-driven perceptual (TDP) loss that facilitates learning to restore task-related features acquired by a task network, enhancing task performance.
    \vspace{2mm}
    \item We propose the cross-quality patch mix and the alternate training framework to address the potential problems of TDP loss, further enhancing the efficacy of TDP loss.
    \vspace{2mm}
    %
    \item The extensive experiments demonstrate that our SR4IR is generally applicable across a wide range of image recognition tasks to enhance task performance while generating perceptually satisfying SR images.
    %
\end{itemize}

\begin{figure*}[t!]
    \centering
    \includegraphics[width=1.0\linewidth]{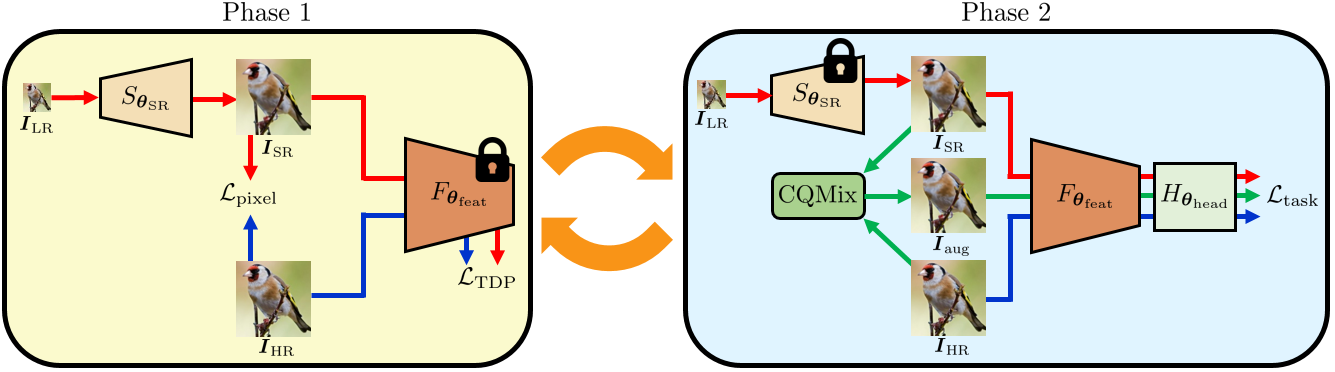}
    \\
    \figspace
    \caption{
        \textbf{Overview of the proposed SR4IR framework.}
        Our SR4IR framework consists of two training phases, where SR and task networks are alternately trained.
        %
        During the first phase, SR4IR updates the SR network using the TDP loss, which is introduced in Section~\ref{subsec:percep}, while the task network is temporarily frozen.
        In the second phase, SR4IR updates the task network using the proposed data augmentation strategy called CQMix, which is introduced in Section~\ref{subsec:mix}, while the SR network is temporarily frozen.
    }
    \label{fig:overview}
    \figxspace
    \vspace{-0.3cm}
\end{figure*}

\section{Related Work}
\Paragraph{Single Image Super-Resolution.}
In the era of deep learning, deep neural network-based methods have risen to prominence in the field of SR.
After the introduction of the simple deep-learning-based SR network SRCNN~\cite{sr_srcnn}, 
VDSR~\cite{sr_vdsr} brought about innovation in network architecture design by stacking deeper convolutional layers with the inclusion of residual connections.
EDSR~\cite{sr_edsr} and SRResNet~\cite{sr_srgan} brought the Residual Block concept into SR networks, and subsequent methods such as RDN~\cite{sr_rdn}, DBPN~\cite{sr_dbpn} further evolved the Dense Residual Block, resulting in high-performance network structures.
%
SAN~\cite{sr_san}, HAN~\cite{sr_han}, and RCAN~\cite{sr_rcan} enhance the SR performance by incorporating channel attention mechanisms, enabling the exploitation of useful channel-wise features.
Recently, transformer-based methods such as SwinIR~\cite{sr_swinir}, Restormer~\cite{db_restormer}, and Uformer~\cite{db_uformer} have demonstrated remarkable SR performance by harnessing the powerful capabilities of Vision Transformers~\cite{misc_vit}.

On the other hand, perceptual super-resolution (SR) methods constitute an alternative approach dedicated to restoring photorealistic SR results.
%
%
Johnson~\etal~\cite{misc_perceptual} introduced perceptual loss, which minimizes the distance in the feature space and has shown its effectiveness in reconstructing high-frequency components in SR.
Zhang~\etal~\cite{zhang2018perceptual} illustrated that the perceptual quality of images is not directly correlated with the peak signal-to-noise ratio (PSNR), then introduced the perceptual metric known as LPIPS.
Subsequent to the development of adversarial loss~\cite{gans}, GAN-based SR methods~\cite{sr_srgan, sr_esrgan, rs_realesrgan, liang2022details, sajjadi2017enhancenet, soh2019natural, park2023content, son2021toward} have been successful in achieving realistic SR results.
%
More recently, diffusion-based SR methods~\cite{saharia2022image, xia2023diffir, yu2024scaling, luo2023image, choi2021ilvr, kawar2022denoising, wang2022zero} have further pushed the boundaries by leveraging a diffusion-based model~\cite{ho2020denoising} to produce even more realistic SR results.

\Paragraph{Image restoration for high-level vision tasks.}
There exist pioneering recent works~\cite{liu2019exploring, cui2022exploring, liu2017image, zhao2018residual, shermeyer2019effects, dai2016image} focused on designing image restoration techniques that offer advantages in high-level computer vision tasks.
Zhou~\etal~\cite{zhou2021improving} introduced a weight map that uses pixel loss primarily in high-frequency regions, showing its effectiveness in image classification.
TDSR~\cite{sr_tdsr} presented an object detection-driven SR method that updates the SR network using a combination of reconstruction and detection losses and provides a comprehensive study on balancing the joint loss.
%
%
Wang~\etal~\cite{Wang2020DSRL} proposed adding an SR branch to the segmentation branch, enabling the effective utilization of HR representation.
Bai~\etal~\cite{bai2018finding, bai2018sod} have demonstrated that incorporating adversarial loss through a multitask generative adversarial network can further enhance the performance of low-quality object detection.
In contrast to previous approaches, our method successfully restores task-specific high-frequency components by leveraging the concept of perceptual loss.
Moreover, we are the first to introduce a generalizable task-combined SR framework that achieves remarkable performance across a range of image recognition tasks.
%

%
\vspace{-0.3cm}
\section{Proposed method}
\label{sec:proposed_method}

\paragraph{Overview.} 
%
%
Figure~\ref{fig:overview} shows the overview of the proposed \textbf{S}uper-\textbf{R}esolution \textbf{f}or \textbf{I}mage \textbf{R}ecognition (SR4IR) framework.
%
The SR4IR comprises two distinct training phases that are alternately executed to train either the super-resolution (SR) or the task network.
%
In the first phase, we train the SR network using the Task-Driven Perceptual~(TDP) loss, which is designed to guide the SR network to restore high-frequency details relevant to the task (Section~\ref{subsec:percep}).
%
In the second phase, we train the task network using the novel Cross-Quality patch Mix~(CQMix) augmentation method, which prevents the task network from learning biased features, further enhancing the effectiveness of the TDP loss (Section~\ref{subsec:mix}).
%
Lastly, we present the effect of alternate training with respect to TDP loss and summarize our final training objectives in each training phase (Section~\ref{subsec:alternate}).

\paragraph{Notation.}
Let $\boldsymbol{I}_\text{HR} \in \mathbb{I}_\text{HR}$ and $\boldsymbol{I}_\text{LR} \in \mathbb{I}_\text{LR}$ denote the high-resolution (HR) and low-resolution (LR) images, paired with their corresponding task labels $\boldsymbol{y} \in \mathbb{Y}$ (\eg, class labels in image classification task).
$S_{\boldsymbol{\theta}_{\text{SR}}}: \mathbb{I}_\text{LR} \rightarrow \mathbb{I}_\text{SR}$ denotes the SR network parameterized by $\boldsymbol{\theta}_{\text{SR}}$ where $\mathbb{I}_\text{SR}$ is the super-resolved image set.
%
Given the SR images~$\boldsymbol{I}_\text{SR}$, we predict the corresponding labels using the task network: $H_{\boldsymbol{\theta}_{\text{head}}} \circ F_{\boldsymbol{\theta}_{\text{feat}}}: \mathbb{I}_\text{SR} \rightarrow \mathbb{Y}$, where $F_{\boldsymbol{\theta}_{\text{feat}}}$ and $H_{\boldsymbol{\theta}_{\text{head}}}$ indicate the feature extractor and the task-specific head module (\eg, the last fully-connected layer for image classification tasks) parameterized by 
$\boldsymbol{\theta}_{\text{feat}}$ and $\boldsymbol{\theta}_{\text{head}}$, respectively.
%

\vspace{-2mm}
\paragraph{Problem Definition.}
The goal of SR4IR is to produce photo-realistic SR images from $\boldsymbol{I}_\text{LR}$ that can benefit the subsequent image recognition task, which is formally given by:
\begin{equation}\label{eq:problem_definition}
    \begin{aligned}
        \min\limits_{\boldsymbol{\theta}_{\text{SR}},\boldsymbol{\theta}_{\text{feat}},\boldsymbol{\theta}_{\text{head}}} & \mathcal{L}_{\text{SR}}(S_{\boldsymbol{\theta}_{\text{SR}}}(\boldsymbol{I}_\text{LR}),\boldsymbol{I}_\text{HR}) \\ & + \mathcal{L}_{\text{task}}(H_{\boldsymbol{\theta}_{\text{head}}} \circ F_{\boldsymbol{\theta}_{\text{feat}}} \circ S_{\boldsymbol{\theta}_{\text{SR}}}(\boldsymbol{I}_\text{LR}),\boldsymbol{y}),
    \end{aligned}
\end{equation}
where the first and second terms are used to evaluate the quality of the super-resolved results (\eg, LPIPS~\cite{zhang2018perceptual}) and the performance for the target high-level vision task (\eg, classification accuracy), respectively.

\subsection{Task-Driven Perceptual Loss}\label{subsec:percep}

High-resolution (HR) images contain high-frequency information that is crucial to solving high-level vision tasks.
However, due to the ill-posed nature of the SR, such task-relevant features are hard to restore using traditional pixel-wise reconstruction loss.
Even with the use of perceptual loss~\cite{misc_perceptual,bruna2016super,leon2015_texture}, designed to produce realistic SR results, there is no guarantee of successfully reconstructing useful features for a specific task.
To address this challenge, we propose task-driven perceptual (TDP) loss that effectively guides the SR network to produce images containing high-frequency details relevant to the task.
Specifically, the TDP loss aims to maximize the similarity between HR and SR images within the latent space of the feature extractor of the task network $F_{\boldsymbol{\theta}_{\text{feat}}}$, whose outputs are substantially relevant to the task.
We define the TDP loss as follows:
\begin{equation}\label{eq:percep}
    \mathcal{L}_{\text{TDP}} = \lVert F_{\boldsymbol{\theta}_\text{feat}}(\boldsymbol{I}_\text{SR}) - F_{\boldsymbol{\theta}_\text{feat}}(\boldsymbol{I}_\text{HR}) \rVert_1.
\end{equation}
%

\vspace{0.3cm}
\subsection{Cross-Quality Patch Mix}\label{subsec:mix}

The effectiveness of TDP loss is inherently related to the training procedure of the task network.
For high-level vision tasks, it is widely known that a network is susceptible to shortcut learning~\cite{geirhos2020shortcut}, which means that a network can rely on biased and non-intrinsic features in training data, leading to poor generalization ability.
In particular, learning biased features can pose challenges in employing the TDP loss, as biased features learned by the task network can hinder the SR network from acquiring comprehensive features present in HR images through the TDP loss.
For instance, if a task network exclusively learns `\textit{beak}' features to classify bird species, the TDP loss may struggle to effectively guide SR images to restore other key high-frequency contents, such as `\textit{wings}' or `\textit{feet}'.
This limitation could hinder the overall task performance boosting that could be attained through the SR method.

\begin{figure}[t!]
    \centering
    \includegraphics[width=1.0\linewidth]{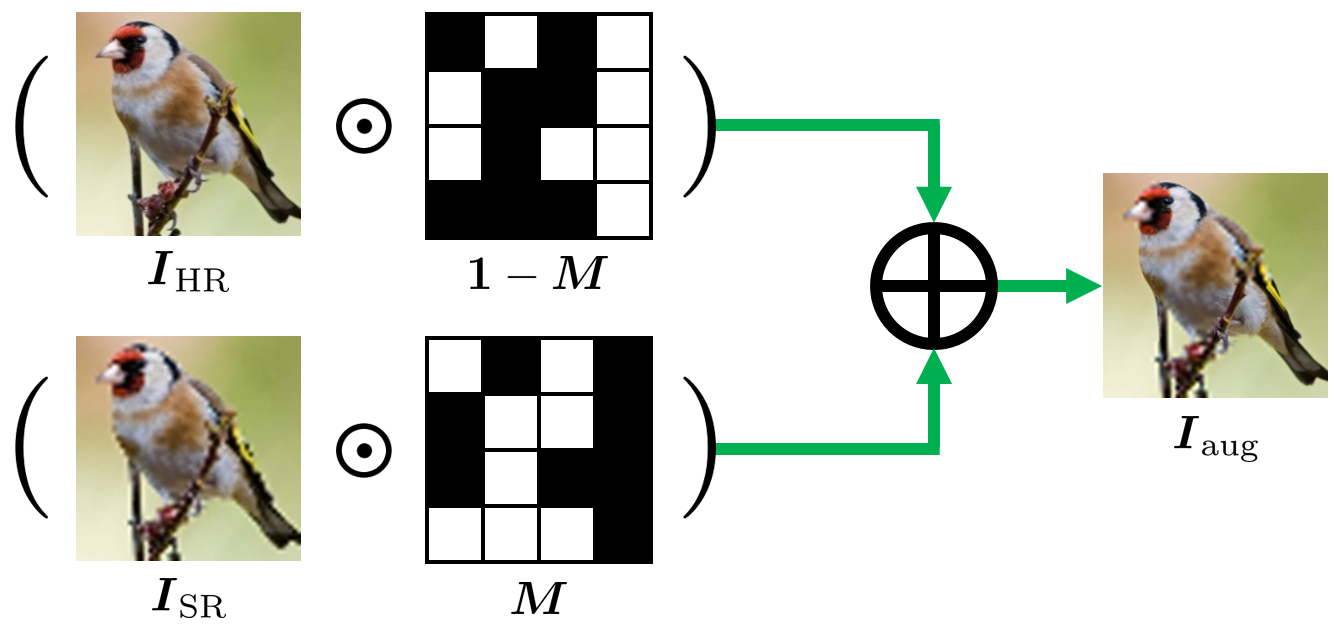}
    \\
    \figspace
    \caption{
        \textbf{Concept of CQMix.}
        The black and white region in the figure represents $\mathbf{0}$ and $\mathbf{1}$ region in the binary mask $\boldsymbol{M}$.
    }
    \label{fig:cqmix}
    \figxspace
    \vspace{-0.0cm}
\end{figure}

To address this issue, we introduce a novel data augmentation strategy called \textbf{C}ross-\textbf{Q}uality patch \textbf{Mix}~(CQMix) when training the task network in the second phase.
%
CQMix is designed to enhance the effectiveness of the TDP loss by encouraging the task network to use diverse high-frequency content as clues.
Specifically, CQMix augments an input image by randomly choosing either HR or SR patches for each grid region, as shown in Figure~\ref{fig:cqmix}, where the number of patches is hyper-parameter.
By doing so, CQMix creates an image in which high-frequency components are randomly erased at a patch level, effectively preventing the task network from learning specific high-frequency features as shortcuts.
Moreover, it is noteworthy that our CQMix is generally applicable to various vision tasks, unlike other data augmentation strategies, such as Mix-Up~\cite{zhang2017mixup} or CutMix~\cite{yun2019cutmix}, as long as HR and LR pairs are available.

\vspace{0.3cm}
\subsection{Alternate Training Framework}\label{subsec:alternate}

Conventional perceptual loss calculates the similarity between SR and HR images within the feature spaces of a \textit{fixed} network, such as fixed VGG~\cite{net_vgg} pre-trained on ImageNet dataset~\cite{data_imagenet}.
However, this approach can be problematic for TDP loss, which aims to restore task-relevant features from HR images.
To effectively utilize the TDP loss to enhance task performance, the task network must be capable of extracting task-relevant features from SR images.
However, since the fixed network is not originally trained on $\boldsymbol{I}_\text{SR}$, which keeps changing during training, it cannot effectively extract task-relevant features from $\boldsymbol{I}_\text{SR}$, thus reducing the effectiveness of the TDP loss.
%
%
To resolve this domain gap issue, we propose an alternative training framework, as illustrated in Figure~\ref{fig:overview}, which enables us to employ the feature space of the \textit{on-training} feature extractor of the task network $F_{\boldsymbol{\theta}_{\text{feat}}}$, for calculating the TDP loss.
By doing so, we can further boost the efficacy of our TDP loss by providing a more meaningful guide to the SR network to acquire task-related knowledge.

Specifically, in the first phase, we train SR network~$S_{\boldsymbol{\theta}_{\text{SR}}}$ with pixel-wise reconstruction loss $\mathcal{L}_{\text{pixel}}$ and the proposed TDP loss, while keeping the feature extractor $F_{\boldsymbol{\theta}_{\text{feat}}}$ frozen.
In the second phase, we train the task network~$H_{\boldsymbol{\theta}_{\text{head}}} \circ F_{\boldsymbol{\theta}_{\text{feat}}}$ with a task-specific loss $\mathcal{L}_{\text{task}}$, \eg, cross-entropy loss for image classification, while freezing the SR network.
We use three types of images in this phase: HR, SR, and augmented images by our CQMix.
In summary, the objectives of our SR4IR in each phase are defined as follows:
\begin{equation}\label{eq:training_loss}
    \begin{aligned}
        \min\limits_{\boldsymbol{\theta}_{\text{SR}}} \:\: & \mathcal{L}_{\text{pixel}}(S_{\boldsymbol{\theta}_{\text{SR}}} (\boldsymbol{I}_{\text{LR}}), \boldsymbol{I}_{\text{HR}}) + \mathcal{L}_{\text{TDP}}, & \text{\textit{in}\,~phase 1,}\\
        \min\limits_{\boldsymbol{\theta}_{\text{feat}},\boldsymbol{\theta}_{\text{head}}} \:\: & \mathcal{L}_{\text{task}}(H_{\boldsymbol{\theta}_{\text{head}}} \circ F_{\boldsymbol{\theta}_{\text{feat}}} (\boldsymbol{I}_{\text{cat}}), \boldsymbol{y}), & \text{\textit{in}\,~phase 2,}\\
    \end{aligned}
\end{equation}
where $\boldsymbol{I}_{\text{cat}}$ is the concatenation of $\boldsymbol{I}_{\text{SR}}$, $\boldsymbol{I}_{\text{HR}}$, and $\boldsymbol{I}_{\text{aug}}$, where the $\boldsymbol{I}_{\text{aug}}$ are the images augmented by the CQMix.

\vspace{0.5cm}
\section{Experiments}
\label{sec:experiments}

%

\paragraph{Datasets.}
%
We select representative datasets and corresponding evaluation metrics for various image recognition tasks, including semantic segmentation, object detection, and image classification.
%
For semantic segmentation and object detection, we use the PASCAL VOC2012 dataset~\cite{pascal-voc-2012}.
%
For the image classification task, we use the Stanford Cars~\cite{krause20133d} and CUB-200-2011~\cite{WahCUB_200_2011} datasets.
Following the conventional SR datasets~\cite{data_div2k, data_set5, data_set14}, we construct the corresponding LR datasets applying bicubic downsampling with scale factors x4 and x8.
%
%
%
%

\paragraph{Evaluation.}
\label{ssec:eval}
We denote the task network $H_{\boldsymbol{\theta}_{\text{head}}} \circ F_{\boldsymbol{\theta}_{\text{feat}}}$ and the SR network $S_{\boldsymbol{\theta}_{\text{SR}}}$ as $T$ and $S$, respectively, for convenience.
Then, the available training scenarios when training both $T$ and $S$ concurrently are as follows:
%
\vspace{3mm}
\begin{itemize}
    \item $\mathbb{I}_\text{HR}\rightarrow{T}\;$:
    Train the $T$ with $\mathcal{L}_\text{task}$ using $\paren{\boldsymbol{I}_\text{HR}, \boldsymbol{y}}$. This can be regarded as an oracle in our problem.
    \vspace{2mm}
    \item $\mathbb{I}_\text{LR}\rightarrow{T}\,$:
    Train the $T$ with $\mathcal{L}_\text{task}$ using $\paren{\mathbf{B}\paren{\boldsymbol{I}_\text{LR}}, \boldsymbol{y}}$, where $\mathbf{B}$ is bilinear upsampling operator.
    \vspace{2mm}
    \item $S\rightarrow{T}\,$:
    Train the $S$ with $\mathcal{L}_\text{pixel}$ using $\paren{\boldsymbol{I}_\text{LR}, \boldsymbol{I}_\text{HR}}$ first.
    Then train $T$ with $\mathcal{L}_\text{task}$ using $\paren{S\paren{\boldsymbol{I}_\text{LR}}, \boldsymbol{y}}$, while freezing $S$, as in previous work~\cite{zhou2021improving}.
    \vspace{2mm}
    \item $T\rightarrow{S}\,$:
    Train $T$ with $\mathcal{L}_\text{task}$ using $\paren{\boldsymbol{I}_\text{HR}, \boldsymbol{y}}$ first. Then, train $S$ with $\mathcal{L}_\text{pixel}+\mathcal{L}_\text{task}$ using $\paren{\boldsymbol{I}_\text{LR}, \boldsymbol{I}_\text{HR}, \boldsymbol{y}}$, while freezing $T$, as in previous work~\cite{sr_tdsr, liu2017image, liu2019exploring}.
    \vspace{2mm}
    \item $S\,+\,T\,$:
    Jointly train both $S$ and $T$ with $\mathcal{L}_\text{pixel}+\mathcal{L}_\text{task}$ using $\paren{\boldsymbol{I}_\text{LR}, \boldsymbol{I}_\text{HR}, \boldsymbol{y}}$.
    \vspace{3mm}
\end{itemize}
%
%
We evaluate the proposed method, named \textbf{S}uper-\textbf{R}esolution \textbf{f}or \textbf{I}mage \textbf{R}ecognition (SR4IR), by comparing it with the above methods.
For evaluation metrics, we use mean Intersection over Union (mIoU) for semantic segmentation, mean Average Precision (mAP) for object detection, and Top-1 accuracy for image classification.
The mAP score is calculated via the COCO~\cite{data_coco} mAP evaluator, which uses 10 IoU thresholds from 0.5 to 0.95 with a step size of 0.05.
%
Furthermore, we use LPIPS~\cite{zhang2018perceptual} and PSNR to measure the quality of the SR images.
%
%

%

\vspace{-3mm}
\paragraph{Network architecture.}
%
For the SR network, we employ two architectures: EDSR-baseline~\cite{sr_edsr} and SwinIR~\cite{sr_swinir}.
%
We initialize the SR network from network weights pre-trained on DIV2K~\cite{data_div2k}, widely used for SR training.
Regarding the task network, we select representative network architectures tailored to each specific task.
%
Specifically, we adopt DeepLabV3~\cite{chen2017rethinking} for semantic segmentation, Faster-RCNN~\cite{det_fasterrcnn} for object detection, and ResNet18~\cite{cls_resnet} for image classification.
The DeepLabV3 and Faster-RCNN are based on MobileNetV3~\cite{cls_mobilenetv3} backbone.
%
The backbone networks are initialized using ImageNet~\cite{data_imagenet} pre-trained weights provided by PyTorch~\cite{fw_pytorch} for easier optimization.

\vspace{-2mm}
\paragraph{Implementations.}
We utilize AdamW~\cite{misc_adamw} and SGD optimizer for the SR and task network.
The learning rates are set to ${10}^{-4}$ for the SR network, $2\times{10}^{-2}$ for DeepLabV3 and Faster-RCNN, and $3\times{10}^{-2}$ for ResNet18, as these values have been shown to produce the best results for their respective tasks.
The batch sizes are set to 16 for both semantic segmentation and object detection and 128 for image classification.
We apply cosine annealing~\cite{misc_cosine_annealing} scheduling to all optimizers, and all experiments are conducted under PyTorch~\cite{fw_pytorch} with eight NVIDIA RTX A6000 GPUs.
For the CQMix, the number of patches is set to 64 for both semantic segmentation and object detection and 16 for image classification, regarding the image resolution for each dataset.

\subsection{Quantitative results} \label{subsec:quantitative}
\paragraph{Semantic segmentation.}
\begin{table*}[t!]
\footnotesize
\centering
\setlength\tabcolsep{1.0pt}
\def\arraystretch{1.1}
\resizebox{0.75\linewidth}{!}{
    \begin{tabular}{L{2.4cm}|C{1.9cm}|C{1.5cm}C{1.5cm}C{1.5cm}|C{1.5cm}C{1.5cm}C{1.5cm}}
    \toprule
    \,~\multirow{2}{*}{Methods} & \multirow{2}{*}{SR Method} &\multicolumn{3}{c|}{x4} & \multicolumn{3}{c}{x8} \\
    \cline{3-8}
    & & mIoU$_\uparrow$ & LPIPS$_\downarrow$ & PSNR$_\uparrow$ & mIoU$_\uparrow$ & LPIPS$_\downarrow$ & PSNR$_\uparrow$ \\
    \hline
    \;~$\mathbb{I}_\text{HR}~{\rightarrow}~T$ & Oracle~($\boldsymbol{I}_\text{HR}$) & 63.3 & 0.000 & $+\inf$ & 63.3 & 0.000 & $+\inf$ \\
    \hline
    \;~$\mathbb{I}_\text{LR}~{\rightarrow}~T$ & Bilinear up & 58.6 & 0.300 & 26.16 & 49.3 & 0.476 & 22.54 \\
    \hline
    \;~$S~{\rightarrow}~T$ & & \underline{60.2} & \underline{0.241} & \textbf{28.94} & \underline{52.3} & \underline{0.407} & \textbf{24.41} \\
    \;~$T~{\rightarrow}~S$ & EDSR- & 56.6 & 0.352 & 26.83 & 49.2 & 0.461 & 23.14 \\
    \;~$S~+~T$ & baseline~\cite{sr_edsr} & 58.0 & 0.318 & 27.64 & 50.0 & 0.480 & 23.58 \\
    \;~\textbf{SR4IR (Ours)} & & \textbf{60.7} & \textbf{0.220} & \underline{28.21} & \textbf{55.0} & \textbf{0.380} & \underline{23.91} \\
    \hline
    \;~$S~{\rightarrow}~T$ & \multirow{4}{*}{SwinIR~\cite{sr_swinir}} & \underline{61.4} & \underline{0.221} & \textbf{29.57} & \underline{53.9} & \underline{0.387} & \textbf{24.75} \\
    \;~$T~{\rightarrow}~S$ & & 53.2 & 0.376 & 26.04 & 48.7 & 0.466 & 23.06 \\
    \;~$S~+~T$ & & 57.5 & 0.341 & 27.55 & 50.0 & 0.480 & 23.63 \\
    \;~\textbf{SR4IR (Ours)} & & \textbf{62.2} & \textbf{0.194} & \underline{28.83} & \textbf{56.5} & \textbf{0.355} & \underline{24.23} \\
    \bottomrule
    \end{tabular}
}
\vspace{-0.1cm}
\caption{
    \textbf{Performance on the semantic segmentation.}
    We adopt DeepLabV3~\cite{chen2017rethinking} with the MobileNetV3~\cite{cls_mobilenetv3} backbone as the task network $T$ and PASCAL VOC2012~\cite{pascal-voc-2012} as an evaluation dataset. The \textbf{bold} and \underline{underlined} mean the best and second best, respectively.
}
\vspace{-0.3cm}
\label{table:semantic_segmentation}
\end{table*}
\begin{table*}[t!]
\footnotesize
\centering
\setlength\tabcolsep{1.0pt}
\def\arraystretch{1.1}
\resizebox{0.75\linewidth}{!}{
    \begin{tabular}{L{2.4cm}|C{1.9cm}|C{1.5cm}C{1.5cm}C{1.5cm}|C{1.5cm}C{1.5cm}C{1.5cm}}
    \toprule
    \,~\multirow{2}{*}{Methods} & \multirow{2}{*}{SR Method} &\multicolumn{3}{c|}{x4} & \multicolumn{3}{c}{x8} \\
    \cline{3-8}
    & & mAP$_\uparrow$ & LPIPS$_\downarrow$ & PSNR$_\uparrow$ &mAP$_\uparrow$ &  LPIPS$_\downarrow$ & PSNR$_\uparrow$ \\
    \hline
    \;~$\mathbb{I}_\text{HR}~{\rightarrow}~T$ & Oracle~($\boldsymbol{I}_\text{HR}$) & 36.7 & 0.000 & $+\inf$ & 36.7 & 0.000 & $+\inf$ \\
    \hline
    \;~$\mathbb{I}_\text{LR}~{\rightarrow}~T$ & Bilinear up & 27.5 & 0.411 & 22.77 & 18.9 & 0.559 & 20.29 \\
    \hline
    \;~$S~{\rightarrow}~T$ & & \underline{30.3} & 0.336 & \textbf{24.62} & \underline{21.9} & 0.494 & \textbf{21.62} \\
    \;~$T~{\rightarrow}~S$ & EDSR- & 27.2 & \underline{0.317} & 23.37 & 15.5 & \underline{0.476} & 20.48 \\
    \;~$S~+~T$ & baseline~\cite{sr_edsr} & 28.3 & 0.354 & 23.71 & 20.3 & 0.506 & \underline{20.98} \\
    \;~\textbf{SR4IR (Ours)} & & \textbf{32.4} & \textbf{0.275} & \underline{23.94} & \textbf{25.5} & \textbf{0.416} & 20.86 \\
    \hline
    \;~$S~{\rightarrow}~T$ & \multirow{4}{*}{SwinIR~\cite{sr_swinir}} & \underline{31.5} & 0.309 & \textbf{25.13} & \underline{23.7} & 0.474 & \textbf{21.85} \\
    \;~$T~{\rightarrow}~S$ & & 29.7 & \underline{0.284} & 23.59 & 18.9 & \underline{0.449} & 20.59 \\
    \;~$S~+~T$ & & 31.4 & 0.331 & 23.94 & 21.9 & 0.489 & \underline{21.11} \\
    \;~\textbf{SR4IR (Ours)} & & \textbf{33.9} & \textbf{0.239} & \underline{24.38} & \textbf{27.8} & \textbf{0.382} & 21.07 \\
    \bottomrule
    \end{tabular}
}
\vspace{-0.1cm}
\caption{
    \textbf{Performance on the object detection.}
    We adopt Faster-RCNN~\cite{det_fasterrcnn} with the MobileNetV3 backbone~\cite{cls_mobilenetv3} as the task network $T$ and PASCAL VOC2012~\cite{pascal-voc-2012} as an evaluation dataset.
    The \textbf{bold} and \underline{underlined} mean the best and second best, respectively.
}
\vspace{-0.3cm}
\label{table:object_detection}
\end{table*}
\begin{table*}[t!]
\small
\centering
\setlength\tabcolsep{1.0pt}
\def\arraystretch{1.1}
\resizebox{0.90\linewidth}{!}{
    \begin{tabular}{L{2.5cm}|C{2.0cm}|C{2.2cm}C{1.5cm}|C{2.2cm}C{1.5cm}|C{2.2cm}C{1.5cm}|C{2.2cm}C{1.5cm}}
    \toprule
    \,~\multirow{3}{*}{Methods} & \multirow{3}{*}{SR Method} & \multicolumn{4}{c|}{StanfordCars~\cite{krause20133d}} & \multicolumn{4}{c}{CUB-200-2011~\cite{WahCUB_200_2011}} \\
    \cline{3-10}
    & & \multicolumn{2}{c|}{x4} & \multicolumn{2}{c|}{x8} & \multicolumn{2}{c|}{x4} & \multicolumn{2}{c}{x8} \\
    \cline{3-10}
    & & Top-1 Acc.$_\uparrow$ (\%) & LPIPS$_\downarrow$ & Top-1 Acc.$_\uparrow$ (\%) & LPIPS$_\downarrow$ & Top-1 Acc.$_\uparrow$ (\%) & LPIPS$_\downarrow$ & Top-1 Acc.$_\uparrow$ (\%) & LPIPS$_\downarrow$ \\
    \hline
    \;~$\mathbb{I}_\text{HR}~{\rightarrow}~T$ & Oracle~($\boldsymbol{I}_\text{HR}$) & 86.4 & 0.000 & 86.4 & 0.000 & 78.3 & 0.000 & 78.3 & 0.000 \\
    \hline
    \;~$\mathbb{I}_\text{LR}~{\rightarrow}~T$ & Bilinear up & 78.7 & 0.304 & 61.0 & 0.478 & 70.8 & 0.298 & 58.2 & 0.465 \\
    \hline
    \;~$S~{\rightarrow}~T$ & & \underline{82.7} & \underline{0.176} & \underline{68.3} & \underline{0.328} & \underline{71.1} & 0.216 & \underline{59.8} & \underline{0.365}  \\
    \;~$T~{\rightarrow}~S$ & EDSR- & 80.1 & 0.190 & 56.9 & 0.448 & 70.5 & \underline{0.208} & 56.5 & 0.401 \\
    \;~$S~+~T$ & baseline~\cite{sr_edsr} & 80.8 & 0.196 & 65.2 & 0.378 & 70.8 & 0.226 & 58.6 & 0.380 \\
    \;~\textbf{SR4IR (Ours)} & & \textbf{83.3} & \textbf{0.158} & \textbf{71.4} & \textbf{0.326} & \textbf{72.8} & \textbf{0.179} & \textbf{62.6} & \textbf{0.326} \\
    \hline
    \;~$S~{\rightarrow}~T$ & \multirow{4}{*}{SwinIR~\cite{sr_swinir}} & \underline{84.0} & \underline{0.159} & \underline{71.4} & \underline{0.308} & 72.8 & 0.201 & \underline{61.2} & \underline{0.349}  \\
    \;~$T~{\rightarrow}~S$ & & 82.0 & 0.162 & 61.0 & 0.397 & 72.1 & \underline{0.190} & 58.7 & 0.377 \\
    \;~$S~+~T$ & & 83.4 & 0.179 & 66.6 & 0.355 & \underline{72.9} & 0.212 & 61.2 & 0.364 \\
    \;~\textbf{SR4IR (Ours)} & & \textbf{85.1} & \textbf{0.136} & \textbf{73.9} & \textbf{0.292} & \textbf{74.7} & \textbf{0.157} & \textbf{65.3} & \textbf{0.301} \\
    \bottomrule
    \end{tabular}
}
\vspace{-0.1cm}
\caption{
    \textbf{Performance on the image classification.}
    We adopt ResNet18~\cite{cls_resnet} with a fully-connected layer as the task network $T$.
    The \textbf{bold} and \underline{underlined} mean the best and second best, respectively.
}
\vspace{-0.6cm}
\label{table:image_classification}
\end{table*}
%

%
Table~\ref{table:semantic_segmentation} shows the quantitative results on the semantic segmentation task.
%
On the x4 SR scale, our SR4IR significantly boosts the mIoU scores compared to the case trained on the bilinear upsampled images~$\paren{\mathbb{I}_\text{LR}\rightarrow{T}}$, by +2.1 and +3.6, for EDSR-baseline and SwinIR, respectively.
%
Remarkably, our method with SwinIR can achieve a mIoU score of $62.2$, which is comparable to the $63.3$ mIoU score achieved by the oracle $\paren{{\mathbb{I}_\text{HR}}\rightarrow{T}}$, demonstrating the effectiveness of combining the SR method when dealing with LR images.
%
Furthermore, SR4IR surpasses the mIoU scores of all SR combined baselines, $S\rightarrow{T}$, $T\rightarrow{S}$, and $S\,+\,T$, with a gain of +0.5 and +2.7 compared to the second-best baselines, in the results of the EDSR-baseline with SR scales of x4 and x8, respectively.
%
This validates that our method successfully restores the high-frequency details that are beneficial for the subsequent task.
%
In particular, compared to ${\mathbb{I}_\text{LR}}\rightarrow{T}$, the performance improvement of the proposed SR4IR becomes even more significant as the SR scale increases from x4 to x8, the performance gain increases from +2.1 to +5.7, for the EDSR-baseline case.
%
This indicates that our method is even more effective when the LR images lose a substantial amount of essential high-frequency details relevant to the task, demonstrating the effectiveness of SR4IR.
%


Furthermore, the proposed SR4IR achieves the best LPIPS score in all cases, with a substantial margin of +12\% gain over the second-best baseline for SwinIR on a scale of x4.
These results demonstrate that our SR results are not only beneficial from a task perspective but are also capable of producing visually appealing results.
Noticeably, separate and sequential training of SR and task networks $\paren{S\rightarrow{T}}$ achieves the best PSNR values but a much lower mIoU score than our SR4IR.
%
The results validate that restoring high-frequency information relevant to the task is much more critical to improving task performance than minimizing the pixel-wise error from ground-truth images.

Moreover, it is noteworthy that training jointly both SR and the task network $\paren{S\,+\,T}$ performs even worse than sequential training of the SR and task networks $\paren{S\rightarrow{T}}$.
This indicates that simultaneously updating both the SR and task networks diminishes the restoring power of the SR due to joint optimization, highlighting the advantages of our alternate training framework.

\vspace{-3mm}
\paragraph{Object detection.}
%
Table~\ref{table:object_detection} presents the numerical results of the object detection.
The proposed SR4IR consistently achieves the best mAP score compared to all SR combined baselines.
Specifically, for the case of x8 SR scale with the SwinIR model, our SR4IR outperforms the task network trained on bilinear upsampled images $\paren{\mathbb{I}_\text{LR}\rightarrow{T}}$ and the second-best SR combined baselines $\paren{S\rightarrow{T}}$ by substantial margins of +8.9 and +4.1, respectively.
%
%

\vspace{-3mm}

\paragraph{Image classification.}
%
Table~\ref{table:image_classification} shows the quantitative results of the image classification task.
Our SR4IR significantly improves image classification performance, remarkably achieving +12.9\% improvement on the x8 scale StanfordCars dataset and a +7.1\% enhancement on the x8 scale CUB-200-2011 dataset when combined with the SwinIR model, compared to the task network trained on bilinear upsampled images $\paren{\mathbb{I}_\text{LR}\rightarrow{T}}$.
Furthermore, we note that our method consistently improves performance across various datasets, demonstrating its strong generalization capability.

\subsection{Qualitative results}
\begin{figure*}[t!]
    \centering
    \captionsetup[subfigure]{labelfont=scriptsize, textfont=scriptsize}
    \renewcommand{\wp}{0.247}
        \subfloat[$\mathbb{I}_\text{LR}~{\rightarrow}~T$]{\includegraphics[width=\wp\linewidth]{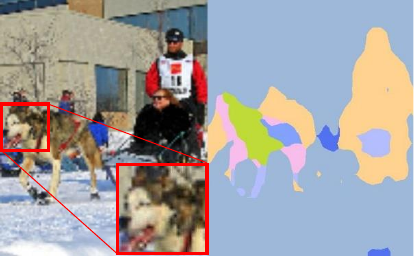}}
        \hfill
        \subfloat[$S~{\rightarrow}~T$]{\includegraphics[width=\wp\linewidth]{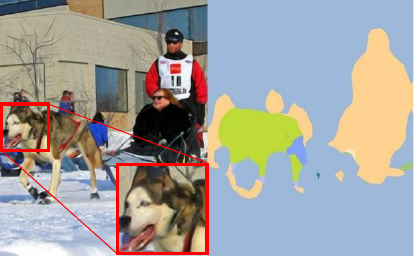}}
        \hfill
        \subfloat[\textbf{SR4IR (Ours)}]{\includegraphics[width=\wp\linewidth]{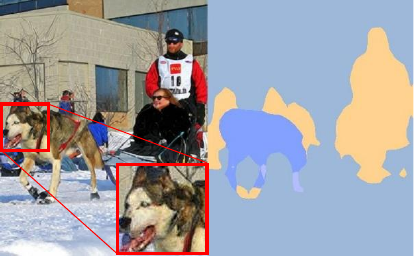}}
        \hfill
        \subfloat[Ground-truth]{\includegraphics[width=\wp\linewidth]{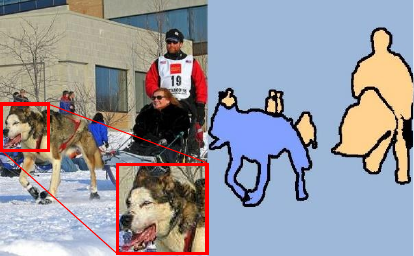}}
    \vspace{-0.3cm}
    \caption{\textbf{Visualization of images and semantic segmentation results on PASCAL VOC dataset.}
        %
        We present the restored images and the corresponding predicted segmentation maps, respectively.
        For (b) and (c), we use the SwinIR model with an SR scale factor of x4.
    }
    \vspace{-0.3cm}
    \label{fig:qualitative_segmentation}
\end{figure*}
\begin{figure*}[t!]
    \centering
    \captionsetup[subfigure]{labelfont=scriptsize, textfont=scriptsize}
    \renewcommand{\wp}{0.247}
        \subfloat[$\mathbb{I}_\text{LR}~{\rightarrow}~T$]{\includegraphics[width=\wp\linewidth]{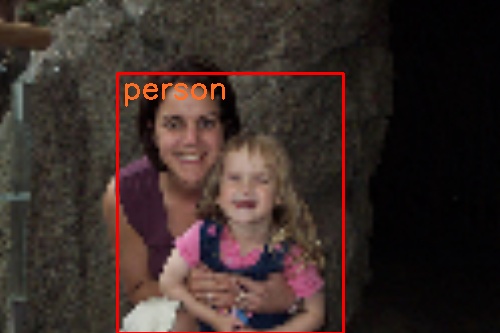}}
        \hfill
        \subfloat[$S~{\rightarrow}~T$]{\includegraphics[width=\wp\linewidth]{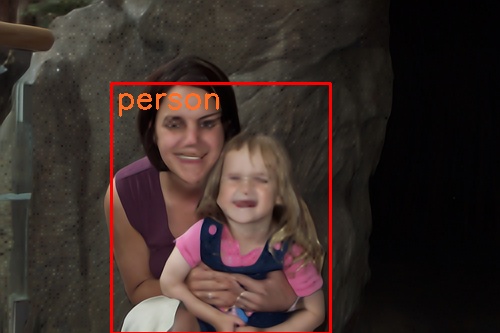}}
        \hfill
        \subfloat[\textbf{SR4IR (Ours)}]{\includegraphics[width=\wp\linewidth]{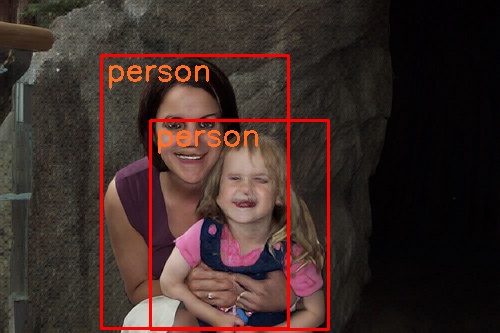}}
        \hfill
        \subfloat[Ground-truth]{\includegraphics[width=\wp\linewidth]{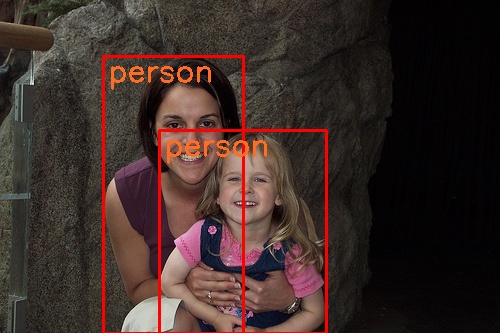}}
    \vspace{-0.3cm}
    \caption{\textbf{Visualization of object detection results on PASCAL VOC dataset.}
        %
        The red box with orange annotation means the predicted object bounding box with the corresponding prediction.
        For (b) and (c), we use the SwinIR model with an SR scale factor of x4.
    }
    \vspace{-0.3cm}
    \label{fig:qualitative_detection}
\end{figure*}
\begin{figure*}[t!]
    \centering
    \captionsetup[subfigure]{labelfont=scriptsize, textfont=scriptsize}
    \renewcommand{\wp}{0.247}
        \subfloat[$\mathbb{I}_\text{LR}~{\rightarrow}~T$]{\includegraphics[width=\wp\linewidth]{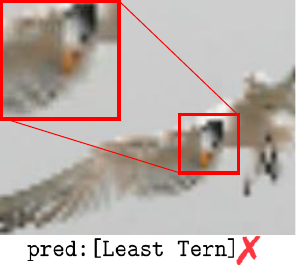}}
        \hfill
        \subfloat[$S~{\rightarrow}~T$]{\includegraphics[width=\wp\linewidth]{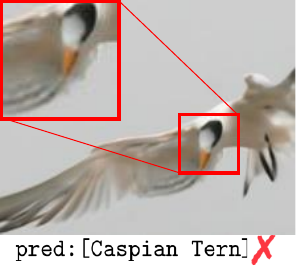}}
        \hfill
        \subfloat[\textbf{SR4IR (Ours)}]{\includegraphics[width=\wp\linewidth]{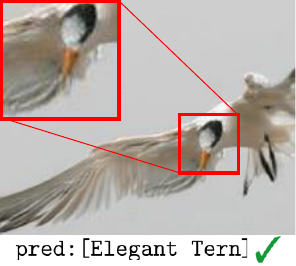}}
        \hfill
        \subfloat[Ground-truth]{\includegraphics[width=\wp\linewidth]{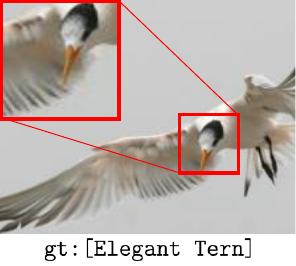}}
    \vspace{-0.3cm}
    \caption{\textbf{Visualization of images and image classification results on CUB-200-2011 dataset.}
        We present the restored images and the corresponding caption.
        The caption below the image represents the predicted image classification results, which is indicated by a checkmark if the prediction is correct.
        For (b) and (c), we use the SwinIR model with an SR scale factor of x4.
    }
    \vspace{-0.4cm}
    \label{fig:qualitative_classification}
\end{figure*}
In this section, we provide visual results of the SR with their corresponding task performance.
Specifically, we compare our SR4IR with two baselines, $\mathbb{I}_\text{LR}\rightarrow{T}$ and $S\rightarrow{T}$, where the latter achieved the second-best quantitative results.
For a more extensive visual comparison with all baselines, please refer to our supplementary materials.

In the context of semantic segmentation, as shown in Figure~\ref{fig:qualitative_segmentation}, the proposed SR4IR successfully generates a segmentation map that closely resembles the ground truth.
In contrast, the baselines $\mathbb{I}_\text{LR}\rightarrow{T}$ and $S\rightarrow{T}$ produce incorrect estimations on `\textit{husky}'.
Regarding object detection, illustrated in Figure~\ref{fig:qualitative_detection}, SR4IR enables the object detection network to accurately identify two people who are closely situated, while $\mathbb{I}_\text{LR}\rightarrow{T}$ and $S\rightarrow{T}$ fail to distinguish and predict them as a single person.
In image classification, as shown in Figure~\ref{fig:qualitative_classification}, the proposed SR4IR empowers the subsequent task network to make the correct label predictions, while the other methods fail to do so.
%
Furthermore, it should be noted that the visualized images produced by our SR4IR achieve the most visually pleasing results.

\subsection{Ablation studies}
In the ablation studies, we analyze the effectiveness of each component of our method and further report the comparison with previous related works.
%
We report the results of semantic segmentation on the x8 scale PASCAL VOC dataset and image classification on the x8 scale StanfordCars dataset, using the EDSR-baseline network.
Please refer to our supplementary materials for more diverse ablation studies.

\vspace{-3mm}
\paragraph{Effectiveness of the TDP loss.}
\begin{table}
\footnotesize
\centering
\setlength\tabcolsep{1.0pt}
\def\arraystretch{1.1}
\resizebox{1.0\linewidth}{!}{
    \begin{tabular}{L{3.5cm}|C{1.2cm}C{1.2cm}|C{2.2cm}C{1.2cm}}
    \toprule
    \,~\multirow{2}{*}{Methods} & \multicolumn{2}{c|}{Segmentation} & \multicolumn{2}{c}{Classification} \\
    \cline{2-5}
    \, & mIoU$_\uparrow$ & LPIPS$_\downarrow$ & Top-1 Acc.$_\uparrow$ (\%) & LPIPS$_\downarrow$ \\
    \hline
    \;~SR4IR \textit{without} TDP loss & 53.3 & 0.407 & 69.9 & 0.327 \\
    \;~SR4IR \textbf{(Ours)} & \textbf{55.0} & \textbf{0.380} & \textbf{71.4} & \textbf{0.326} \\
    \bottomrule
    \end{tabular}
}
\vspace{-0.3cm}
\caption{
    \textbf{Effectiveness of the proposed TDP loss.}
    We evaluate the impact of TDP loss by including or excluding $\mathcal{L}_\text{TDP}$ in phase 1 of equation~\eqref{eq:training_loss} on the performance of SR4IR.
}
\vspace{-0.5cm}
\label{table:ablation_tdp}
\end{table}
Table~\ref{table:ablation_tdp} shows the task performance of our SR4IR, comparing the cases with and without the proposed TDP loss.
%
Without the TDP loss, the performance of SR4IR decreases significantly, with a decrease of -1.7 mIoU score in semantic segmentation and a decrease in precision of -1.5\% in image classification.
The results demonstrate that the TDP loss is crucial for a significant boost in task performance in our SR4IR, showing that restoring the task-relevant high-frequency details is essential in enhancing the performance of the subsequent task network.
Furthermore, the absence of TDP loss results in a significant deterioration in the LPIPS score with a -7\% decrease in segmentation.
These results validate that our TDP loss not only enhances task performance but also contributes to generating perceptually realistic SR results.

\vspace{-3mm}
\paragraph{Effectiveness of the CQMix.}
To verify the effectiveness of the proposed CQMix, we conduct experiments on training the task network with various types of images, as shown in Table~\ref{table:ablation_cqmix}.
When we train the task network with only $\boldsymbol{I}_\text{SR}$, the performance is poor, indicating that the effectiveness of the TDP loss decreases when the task network is not trained to properly utilize the high-frequency information present in HR images.
%
%
However, training the task network solely on $\boldsymbol{I}_\text{HR}$ also yielded subpar results, as the task network is not trained to handle $\boldsymbol{I}_\text{SR}$.
Combining both $\boldsymbol{I}_\text{SR}$ and $\boldsymbol{I}_\text{HR}$ performs better compared to individual methods.
Finally, adding augmented images through our CQMix further enhances the performance, with gains of +0.3 in segmentation and +1.1\% in classification, respectively, highlighting the substantial benefits of preventing shortcut learning.

\vspace{-4mm}
\paragraph{Effectiveness of the alternate training framework.}
To validate the effectiveness of the proposed alternate training framework, we compare our methods with three perceptual-loss-related approaches, denoted as \textit{(A)}, \textit{(B)} and \textit{(C)}.
%
In \textit{(A)}, we substitute the proposed TDP loss to the conventional perceptual loss, which uses a fixed VGG network pre-trained on ImageNet.
%
In \textit{(B)}, rather than using the on-training network proposed in our SR4IR, we employ TDP loss using the fixed task network pre-trained on the corresponding target task, \ie, $\mathbb{I}_\text{HR}~{\rightarrow}~T$.
In \textit{(C)}, rather than adopting alternate training, we apply the TDP loss with a trainable on-training task network within a joint training, where both the SR and task networks are trained jointly.
%

Table~\ref{table:ablation_alternate} presents performance results between (\textit{A}), (\textit{B}), (\textit{C}), and our method.
%
Compared to the performance of SR4IR without TDP loss, which achieves a 53.3 mIoU score and 69.9\% accuracy as presented in Table~\ref{table:ablation_tdp}, the method \textit{(A)} achieves similar or even worse performance, demonstrating that adopting conventional perceptual loss does not help to boost the task performance.
The method \textit{(B)} also results in marginal performance gains, as the fixed and pre-trained task network is not trained to handle the SR results.
%
Notably, the method \textit{(C)} fails to learn because optimizing the TDP loss leads to undesirable local optima when the task network is trainable, which results in forcing the task network to produce a constant output regardless of varied inputs.
In contrast, our SR4IR achieves significant performance improvement compared to the above approaches, highlighting the importance of adopting an alternate training framework to fully leverage the power of the TDP loss.

\vspace{-4mm}
\paragraph{Comparison with previous methods.}
To highlight the advantages of our SR4IR, we include a quantitative comparison with two related previous methods, TDSR~\cite{sr_tdsr} and SOD-MTGAN~\cite{bai2018sod}, which are task-driven restoration methods.
The results are presented in Table~\ref{table:previous}.
Due to the absence of a publicly available training code, we report the results of our re-implemented version, which is elaborated in the supplementary material.
%
In comparison to TDSR and SOD-MTGAN, our SR4IR achieves significantly superior performance, underlining the effectiveness of leveraging perceptual loss when compared to the case of training without perceptual loss (TDSR) and adversarial loss (SOD-MTGAN).

\begin{table}[t!]
\footnotesize
\centering
\setlength\tabcolsep{1.0pt}
\def\arraystretch{1.1}
\resizebox{0.9\linewidth}{!}{
    \begin{tabular}{L{3.8cm}|C{1.8cm}|C{2.2cm}}
    \toprule
    \,~\multirow{2}{*}{Training image types} & Segmentation & Classification \\
    & (mIoU$_\uparrow$) & (Top-1 Acc.\%$_\uparrow$) \\
    \hline
    \;~$\boldsymbol{I}_\text{SR}$ & 51.7 & 66.7 \\
    \;~$\boldsymbol{I}_\text{HR}$ & 53.3 & 59.3 \\
    \;~$\boldsymbol{I}_\text{SR}+\boldsymbol{I}_\text{HR}$ & 54.7 & 70.3  \\
    \;~$\boldsymbol{I}_\text{SR}+\boldsymbol{I}_\text{HR}+\boldsymbol{I}_\text{aug}$~\textbf{(Ours)} & \textbf{55.0} & \textbf{71.4} \\
    \bottomrule
    \end{tabular}
}
\vspace{-2mm}
\caption{
    \textbf{Effectiveness of the proposed CQMix.}
    We compare the performance of SR4IR by varying the training images $\boldsymbol{I}_{\text{cat}}$ in phase 2 of equation~\eqref{eq:training_loss}.
}
\vspace{-0.3cm}
\label{table:ablation_cqmix}
\end{table}
\begin{table}[t!]
\footnotesize
\centering
\setlength\tabcolsep{1.0pt}
\def\arraystretch{1.1}
\resizebox{1.0\linewidth}{!}{
    \begin{tabular}{L{1.2cm}|L{4.2cm}|C{1.7cm}|C{2.0cm}}
    \toprule
    \,~\multirow{2}{*}{Methods} & \;\multirow{2}{*}{Feature extractor for perceptual loss} & Segmentation & Classification \\
    & & (mIoU$_\uparrow$) & (Top-1 Acc.\%$_\uparrow$) \\
    \hline
    \;~\textit{(A)} & \;Pre-trained on ImageNet & 53.6 & 69.4 \\
    \;~\textit{(B)} & \;Pre-trained on corresponding task & 53.6 & 70.3 \\
    \;~\textit{(C)} & \;On-training (joint) & 3.5 & 3.7 \\
    \;~\textbf{Ours} & \;On-training (alternate) & \textbf{55.0} & \textbf{71.4} \\
    \bottomrule
    \end{tabular}
}
\vspace{-2mm}
\caption{
    \textbf{Effectiveness of the proposed alternate training framework.}
    We compare our SR4IR with three perceptual-loss-related approaches, by varying the feature extractor $F_{\boldsymbol{\theta}_\text{feat}}$ used for the TDP loss in equation~\eqref{eq:percep}.
}
\vspace{-0.3cm}
\label{table:ablation_alternate}
\end{table}
\begin{table}[t!]
\footnotesize
\centering
\setlength\tabcolsep{1.0pt}
\def\arraystretch{1.1}
\resizebox{0.8\linewidth}{!}{
    \begin{tabular}{L{3.0cm}|C{1.7cm}|C{2.2cm}}
    \toprule
    \,~\multirow{2}{*}{Methods} & Segmentation & Classification \\
    & (mIoU$_\uparrow$) & (Top-1 Acc. \%$_\uparrow$) \\
    \hline
    \;~TDSR$\dag$~\cite{sr_tdsr} & 49.8 & 56.4 \\
    \;~SOD-MTGAN$\dag$~\cite{bai2018sod} & 51.6 & 68.1 \\    
    \;~\textbf{SR4IR (Ours)} & \textbf{55.0} & \textbf{71.4} \\
    \bottomrule
    \end{tabular}
}
\vspace{-2mm}
\caption{
    \textbf{Comparison with related previous methods.}
    The symbol $\dag$ denotes the results of our re-implementation version.
}
\vspace{-0.5cm}
\label{table:previous}
\end{table}

\section{Conclusion}
In this paper, we tackle the practical scenarios for image recognition tasks where low-resolution (LR) content prevails when processing the images.
We tackle the challenge by employing image super-resolution (SR) techniques.
Considering the ill-posed nature of SR, we focus on developing methods that encourage an SR network to reconstruct task-related features beneficial for a specific task.
In particular, we propose task-driven perceptual (TDP) loss that effectively guides the restoration of task-related high-frequency contents.
Furthermore, our cross-quality patch mix (CQMix) and alternate training framework help us fully harness the advantages of our TDP loss by preventing a task network from learning biased features and eliminating the domain gap problem.
Extensive experimental results demonstrate that our \textbf{S}uper-\textbf{R}esolution \textbf{f}or \textbf{I}mage \textbf{R}ecognition (SR4IR) can be generally and effectively applicable to various image recognition tasks, such as semantic segmentation, object detection, and image classification.

\vspace{3mm}
\noindent\textbf{Acknowledgments.}
This work was supported in part by the IITP grants [No. 2021-0-01343, Artificial Intelligence Graduate School Program (Seoul National University), No.2021-0-02068, and No.2023-0-00156], the NRF grant [No.2021M3A9E4080782] funded by the Korean government (MSIT).

{
    \small
    \bibliographystyle{ieeenat_fullname}
    \bibliography{egbib}
}

\newpage
\
\newpage

\setcounter{section}{0}
\setcounter{figure}{0}
\setcounter{table}{0}
\setcounter{equation}{0}

\renewcommand{\thetable}{S\arabic{table}}
\renewcommand{\thesection}{S\arabic{section}}
\renewcommand{\thefigure}{S\arabic{figure}}
\renewcommand{\theequation}{S\arabic{equation}}

\begin{figure*}[t!]
    \centering
    \captionsetup[subfigure]{labelfont=scriptsize, textfont=scriptsize}
    \renewcommand{\wp}{0.33}
        \subfloat[$\boldsymbol{I}_\text{HR} \leftrightarrow \boldsymbol{I}_\text{SR(pixel)}$]{\includegraphics[width=\wp\linewidth]{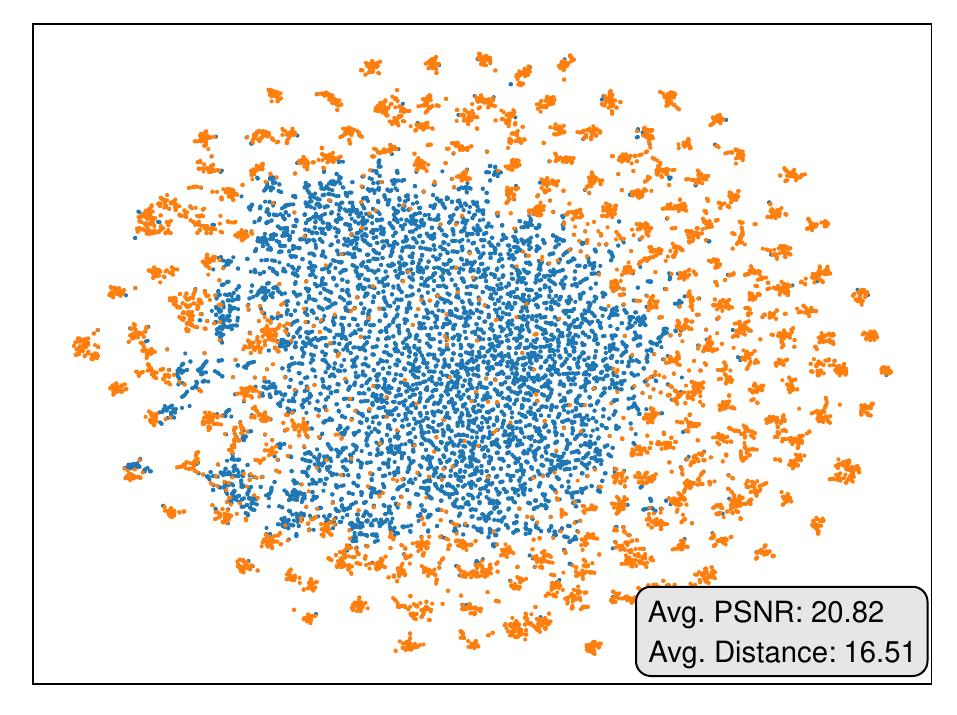}}
        \hfill
        \subfloat[$\boldsymbol{I}_\text{HR} \leftrightarrow \boldsymbol{I}_\text{SR(pixel+percep)}$]{\includegraphics[width=\wp\linewidth]{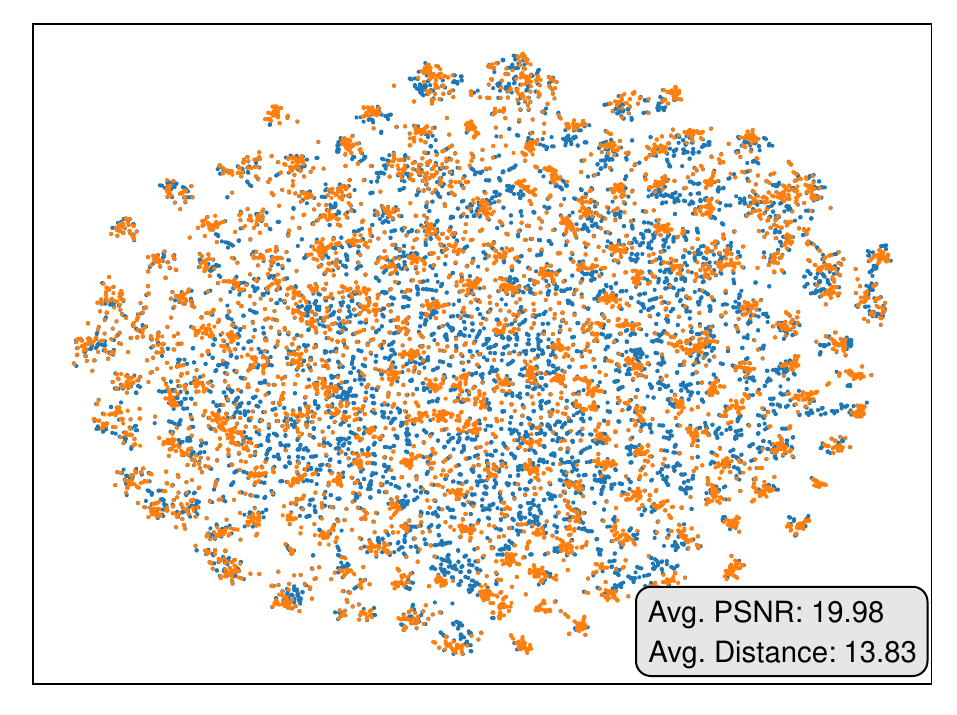}}
        \hfill
        \subfloat[$\boldsymbol{I}_\text{HR} \leftrightarrow \boldsymbol{I}_\text{SR(SR4IR)}$ \textbf{(Ours)}]{\includegraphics[width=\wp\linewidth]{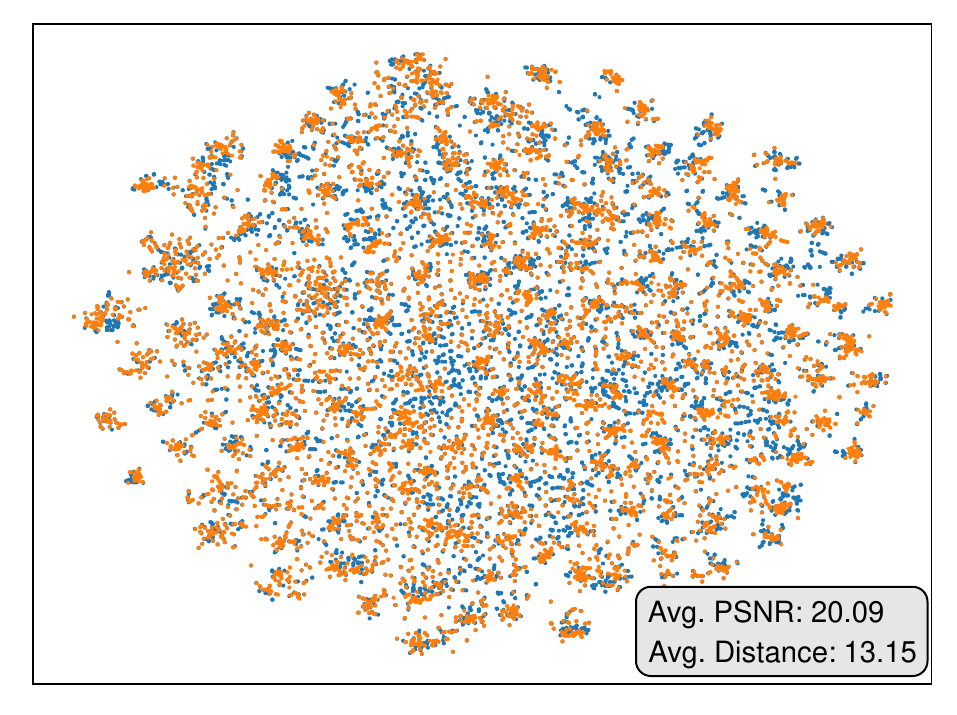}}
    \vspace{-0.2cm}
    \caption{\textbf{t-SNE visualization results.}
        We visualize the t-SNE plots based on the final feature of the feature extractor of a task network trained with HR images.
        The \textcolor{BurntOrange}{orange} and \textcolor{DarkBlue}{blue} points represent the HR and SR samples, respectively.
        We display the average PSNR in the RGB space and the average Euclidean distance in the feature space between the HR and SR pairs, in the bottom right corner of each figure.
        For the distance, the lower value indicates a higher resemblance between the SR results and the HR counterpart.
    }
    \vspace{-0.5cm}
    \label{fig:tsne}
\end{figure*}
In this supplementary document, we show the additional results and ablation studies omitted from the main manuscript due to the lack of space and describe the details:
\begin{itemize}
    \item \ref{sec:effectiveness_sr}. Effectiveness of our SR results
    \item \ref{sec:efficiency}. Efficiency of our SR4IR
    \item \ref{sec:analysis_cqmix}. Analysis on CQMix
    \item \ref{sec:effectiveness_task_driven}. Effectiveness of the task-driven training
    \item \ref{sec:diverse_degradation}. Diverse degradation scenarios for SR4IR
    \item \ref{sec:benchmark}. Evaluation on the SR benchmark dataset
    \item \ref{sec:training_details}. Training details
    \item \ref{sec:previous_details}. Details for reproducing the previous works
    \item \ref{sec:more_visualization}. More visualization comparisons 
\end{itemize}

\section{Effectiveness of our SR results}
\label{sec:effectiveness_sr}
In the main manuscript, we argue that our SR4IR can restore task-relevant high-frequency details that are beneficial for a subsequent image recognition task.
To further demonstrate our claim, we compare the SR results trained by our SR4IR (referred to as $\boldsymbol{I}_\text{SR(SR4IR)}$) with two other sets of SR results that are trained by (1) using pixel-wise reconstruction loss (referred to as $\boldsymbol{I}_\text{SR(pixel)}$) and (2) employing a combination of pixel-wise loss and conventional perceptual loss (referred to as $\boldsymbol{I}_\text{SR(pixel+percep)}$).

\vspace{-0.3cm}
\paragraph{Performance on $S\rightarrow{T}$ setting.} \label{sec:s2t}
We compare the performance of the task network trained by the three types of SR images following the same setting as $S\rightarrow{T}$ described in Section~\ref{ssec:eval}.
Table~\ref{table:s2t} shows that using $\boldsymbol{I}_\text{SR(SR4IR)}$ achieves significantly superior performance (71.1\%) compared to the cases using $\boldsymbol{I}_\text{SR(pixel)}$ (68.3\%) and $\boldsymbol{I}_\text{SR(pixel+percep)}$ (69.7\%).
Furthermore, it is worth noting that the $S\rightarrow{T}$ setting with $\boldsymbol{I}_\text{SR(SR4IR)}$ achieves performance comparable to our final performance, which attains the accuracy of 71.4\%, as presented in Table~\ref{table:image_classification}.
%
These results demonstrate that the performance improvement in our SR4IR model is largely attributed to the SR results, in which the task-relevant high-frequency details are successfully restored.

\vspace{-0.3cm}
\paragraph{Performance on task network trained on HR.} \label{sec:tsne}
To further investigate the similarity between HR and SR images in terms of task-relevant features, we evaluate the SR results $\boldsymbol{I}_\text{SR(pixel)}$, $\boldsymbol{I}_\text{SR(pixel+percep)}$, and $\boldsymbol{I}_\text{SR(SR4IR)}$ using a task network that has been exclusively trained on HR images (referred to as $T_\text{HR}$).
Figure~\ref{fig:tsne} shows the t-SNE~\cite{van2008visualizing} visualization results of each SR result within the feature space of $T_\text{HR}$.
Our $\boldsymbol{I}_\text{SR(SR4IR)}$ exhibits the closest resemblance to HR images in the feature space of $T_\text{HR}$ compared to $\boldsymbol{I}_\text{SR(pixel)}$ and $\boldsymbol{I}_\text{SR(pixel+percep)}$.
In addition, it should be noted that $\boldsymbol{I}_\text{SR(pixel)}$ resembles poorly HR images even with the highest PSNR values, indicating that the PSNR is a less important factor in representing task-relevant features.
Table~\ref{table:h2t} further shows the superior performance of our $\boldsymbol{I}_\text{SR(SR4IR)}$ when the SR results are evaluated by $T_\text{HR}$.
These results demonstrate that $\boldsymbol{I}_\text{SR(SR4IR)}$ contains high-frequency contents closely related to tasks, validating the effectiveness of our SR4IR.

\begin{table}[!t]
\footnotesize
\centering
\setlength\tabcolsep{1.0pt}
\def\arraystretch{1.1}
\resizebox{0.75\linewidth}{!}{
    \begin{tabular}{L{3.0cm}|C{2.2cm}}
    \toprule
    \,~Training images & Top-1 Acc.$_\uparrow$ (\%) \\
    \hline
    \;~$\boldsymbol{I}_\text{SR(pixel)}$ & 68.3 \\
    \;~$\boldsymbol{I}_\text{SR(pixel+percep)}$ & 69.7 \\    
    \;~$\boldsymbol{I}_\text{SR(SR4IR)}$~\textbf{(Ours)} & \textbf{71.1} \\
    \bottomrule
    \end{tabular}
}
\vspace{-2mm}
\caption{
    \textbf{Performance on $S\rightarrow{T}$ setting.}
    We evaluate the image classification accuracy on the StandfordCars dataset with an SR scale of x8.
    We use the EDSR-baseline as an SR network.
}
\vspace{-2mm}
\label{table:s2t}
\end{table}
\begin{table}[!t]
\footnotesize
\centering
\setlength\tabcolsep{1.0pt}
\def\arraystretch{1.1}
\resizebox{0.75\linewidth}{!}{
    \begin{tabular}{L{3.2cm}|C{2.2cm}}
    \toprule
    \,~Test images & Top-1 Acc.$_\uparrow$ (\%) \\
    \hline
    \;~$\boldsymbol{I}_\text{HR}$ (Oracle) & 86.4 \\
    \hline
    \;~$\boldsymbol{I}_\text{SR(pixel)}$ & 29.4 \\
    \;~$\boldsymbol{I}_\text{SR(pixel+percep)}$ & 46.5 \\    
    \;~$\boldsymbol{I}_\text{SR(SR4IR)}$~\textbf{(Ours)} & \textbf{52.3} \\
    \bottomrule
    \end{tabular}
}
\vspace{-2mm}
\caption{
    \textbf{Performance on the task network trained with HR.}
    We evaluate the image classification accuracy on the StandfordCars dataset with an SR scale of x8.
}
\vspace{-2mm}
\label{table:h2t}
\end{table}
\begin{table}[!t]
\footnotesize
\centering
\setlength\tabcolsep{1.0pt}
\def\arraystretch{1.1}
\resizebox{1.0\linewidth}{!}{
    \begin{tabular}{L{2.0cm}|C{1.0cm}|C{1.2cm}C{1.2cm}C{1.2cm}C{1.6cm}C{1.5cm}}
    \toprule
    \, & mIoU$_\uparrow$ & \# Params & GFLOPs & Memory Cache & Training time (GPU hours) & Throughput (img/s) \\
    \hline
    \;~$\mathbb{I}_\text{LR}~{\rightarrow}~T$ & 49.3 & 11.0M & 17.5 & 456MB & 0.64 & 33.8 \\
    \;~$\mathbb{I}_\text{LR}~{\rightarrow}~T_\text{Large}$ & 50.2 & 42.0M & 305.5 & 1621MB& 4.25 & 15.2 \\
    \;~SR4IR \textbf{(Ours)} & \textbf{55.0} & 12.7M & 49.5 & 695MB & 2.41 & 25.3 \\
    \bottomrule
    \end{tabular}
}
\vspace{-2mm}
\caption{
    \textbf{Efficiency of our SR4IR on x8 segmentation task.}
    GFLOPs are calculated based on 480x480 resolution images.
    Memory cache and throughput are evaluated using the PASCAL VOC~\cite{pascal-voc-2012} validation set and an NVIDIA RTX A6000 GPU.
    We use~\cite{sr_edsr} as an SR network.
    The $T$ and $T_\text{Large}$ are DeepLabV3~\cite{chen2017rethinking} with MobileNetV3~\cite{cls_mobilenetv3} and ResNet50~\cite{cls_resnet} backbone.
}
\vspace{-5mm}
\label{table:efficiecny}
\end{table}
\section{Efficiency of our SR4IR}
\label{sec:efficiency}

In Table~\ref{table:efficiecny}, we assess the efficiency of our SR4IR with the baseline method  $\mathbb{I}_\text{LR}\rightarrow{T}$.
As introduced in Section~\ref{ssec:eval} of the main manuscript, the baseline method trains a task network using bilinear-upscaled LR images without the assistance of SR networks.
Despite the superior task performance, our SR4IR incurs a certain degree of computational cost increase compared to $\mathbb{I}_\text{LR}\rightarrow{T}$ due to introducing the SR network.
Nevertheless, we highlight that the performance gain from SR4IR does not simply come from the increased computational costs, demonstrated through a comparison with the baseline using a larger backbone model, $\mathbb{I}_\text{LR}\rightarrow{T_\text{Large}}$. 
In terms of training time, our SR4IR does require some additional training time compared to $\mathbb{I}_\text{LR}\rightarrow{T}$ due to the incorporation of the SR network.
%
For the real-time capability, we observe that the throughput of our SR4IR experiences a modest decrease compared to $\mathbb{I}_\text{LR}\rightarrow{T}$ but still faster than $\mathbb{I}_\text{LR}\rightarrow{T_\text{Large}}$, indicating practical applicability of SR4IR.
\section{Analysis on CQMix}
\label{sec:analysis_cqmix}

\begin{figure}[!t]
    \centering
    \includegraphics[width=0.32\linewidth]{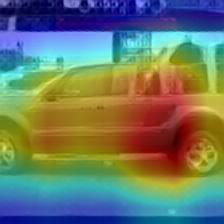}
    \hfill
    \includegraphics[width=0.32\linewidth]{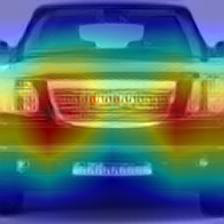}
    \hfill
    \includegraphics[width=0.32\linewidth]{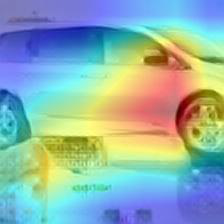}
    \\
    \includegraphics[width=0.32\linewidth]{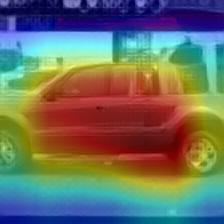}
    \hfill
    \includegraphics[width=0.32\linewidth]{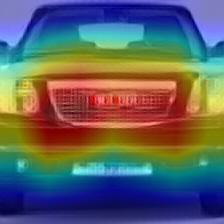}
    \hfill
    \includegraphics[width=0.32\linewidth]{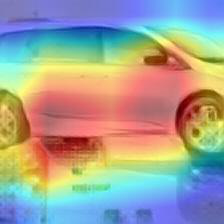}
    \\
    \figspace
    \caption{
        \textbf{Grad-CAM visualization results according to the CQMix.}
        The upper and lower rows represent the Grad-CAM~\cite{selvaraju2017grad} results from the task network trained without and with the proposed CQMix.
        We use EDSR-baseline~\cite{sr_edsr} with a scale factor of x8 and StandfordCars~\cite{krause20133d} dataset.
    }
    \label{fig:gradcam}
    \vspace{-3mm}
\end{figure}

In Section~\ref{subsec:mix}, we propose the CQMix to prevent a task network from learning biased features that could undermine the effectiveness of TDP loss.
%
To further analyze the effect of the CQMix, we visualize class-activation maps using Grad-CAM~\cite{selvaraju2017grad} obtained from a task network trained by our framework with/without the CQMix, which are presented in Figure~\ref{fig:gradcam}.
%
When CQMix is not used (upper row), the task network tends to focus on specific image features, the \ie shortcut feature~\cite{geirhos2020shortcut}, such as car wheels or headlights.
%
In contrast, when CQMix is used (lower row), the task network tends to focus on wider regions and utilizes additional diverse image features, such as the car body and emblem.
%
These visualization results demonstrate that our CQMix is effective in preventing the task network from learning shortcut features, leading to performance improvements when combined with the TDP loss in our framework.

\section{Effectiveness of the task-driven training}
\label{sec:effectiveness_task_driven}

In Table~\ref{table:task_driven}, we show the effectiveness of task-driven training by evaluating the performance across three image recognition tasks with SR networks specially tailored for each task.
Compared to the case of `task-driven' SR, which is presented in the \textcolor{gray}{\textbf{diagonal}} entries, using the SR network trained for different tasks largely degrades the performance.
These results demonstrate the importance of employing a task-driven SR network in improving task performance.

\section{Diverse degradation scenarios for SR4IR}
\label{sec:diverse_degradation}

Table~\ref{table:diverse_degradation} shows the performance of our SR4IR on diverse versions of degraded LR images, which have undergone bicubic downsampling with an additional Gaussian blur filter.
Our SR4IR consistently improves the task performance in all degradation cases, indicating its general applicability.
%
Moreover, the performance gain from SR4IR becomes more pronounced as the degradation severity increases.
This observation is consistent with the discussions in Section~\ref{subsec:quantitative} of our main manuscript, where we noted that the performance improvement of SR4IR is more significant as the SR scale changes from x4 to x8.

\begin{table}[!t]
\footnotesize
\centering
\setlength\tabcolsep{1.0pt}
\def\arraystretch{1.1}
\resizebox{1.0\linewidth}{!}{
    \begin{tabular}{L{2.2cm}|C{2.0cm}C{2.0cm}C{2.0cm}}
    \toprule
    \;~\multirow{2}{*}{Used SR network} & Segmentation & Detection & Classification \\
    \, & (mIoU$_\uparrow$) & (mAP$_\uparrow$) & (Top1 Acc.$_\uparrow$) \\
    \hline
    \;~$S_\text{Segmentation}$ & \cellcolor{gray!25}\textbf{55.0} & 22.4 & 61.0 \\
    \;~$S_\text{Detection}$ & 50.3 & \cellcolor{gray!25}\textbf{25.5} & 60.2 \\
    \;~$S_\text{Classification}$ & 48.4 & 20.0 & \cellcolor{gray!25}\textbf{71.4} \\
    \bottomrule
    \end{tabular}
}
\vspace{-2mm}
\caption{
    \textbf{SR4IR performance according to the SR network.}
    $S_\textit{Task}$ represents the SR network specifically tailored for each \textit{Task} through our SR4IR.
    We use EDSR-baseline~\cite{sr_edsr}, PASCAL VOC~\cite{pascal-voc-2012} and StanfordCars~\cite{krause20133d} with x8 downsampling.
}
\vspace{-0.5cm}
\label{table:task_driven}
\end{table}
%
\begin{table}[!t]
\footnotesize
\centering
\setlength\tabcolsep{1.0pt}
\def\arraystretch{1.1}
\resizebox{1.0\linewidth}{!}{
    \begin{tabular}{L{1.8cm}|C{1.0cm}C{1.0cm}|C{1.0cm}C{1.0cm}|C{1.0cm}C{1.0cm}|C{1.0cm}C{1.0cm}}
    \toprule
    \, & \multicolumn{2}{c|}{\multirow{2}{*}{Bicubic}} & \multicolumn{6}{c}{Gaussian blur~+~Bicubic} \\
    \cline{4-9}
    \, & & & \multicolumn{2}{c|}{std~=~5.0} & \multicolumn{2}{c|}{std~=~10.0} & \multicolumn{2}{c}{std~=~15.0} \\
    \cline{2-9}
    \, & mIoU$_\uparrow$ & LPIPS$_\downarrow$ & mIoU$_\uparrow$ & LPIPS$_\downarrow$ & mIoU$_\uparrow$ & LPIPS$_\downarrow$ & mIoU$_\uparrow$ & LPIPS$_\downarrow$ \\
    \hline
    \;~$\mathbb{I}_\text{LR}~{\rightarrow}~T$ & 49.3 & 0.476 & 47.9 & 0.543 & 38.3 & 0.631 & 29.5 & 0.650 \\
    \;~SR4IR (\textbf{Ours}) & \textbf{55.0} & \textbf{0.380} & \textbf{56.4} & \textbf{0.367} & \textbf{49.8} &\textbf{ 0.48}5 & \textbf{41.4} & \textbf{0.543} \\
    \bottomrule
    \end{tabular}
}
\vspace{-2mm}
\caption{
    \textbf{Diverse degradation scenarios on x8 segmentation.}
    We use the EDSR-baseline~\cite{sr_edsr} as an SR network.
    We evaluate the performance on the PASCAL VOC~\cite{pascal-voc-2012} dataset.
}
\vspace{-0.2cm}
\label{table:diverse_degradation}
\end{table}
\section{Evaluation on the SR benchmark datasets}
\label{sec:benchmark}
Table~\ref{table:benchmark} shows the performance of our task-driven SR network on SR benchmark datasets.
Compared to conventional SR methods trained on pixel-wise loss, the SR network trained by our SR4IR framework exhibits lower PSNR/SSIM values on SR benchmark datasets.
However, as discussed in Section~\ref{sec:effectiveness_sr} and \ref{subsec:quantitative} of our main manuscript, we highlight that restoring task-relevant high-frequency details is crucial for the task performance rather than such distortion-oriented metrics.
\section{Training details}
\label{sec:training_details}
The number of training epochs is set to 100 for semantic segmentation, 30 for object detection, and 200 for image classification.
The task loss is set to cross-entropy loss in semantic segmentation and image classification, and a combination of classification, roi box regression, objectness, and rpn box regression loss in object detection, following the official PyTorch implementation github~\cite{torchvision2016}.
In the case of applying the TDP loss, we exclude the TDP loss during the initial one-tenth of the training process to allow the task network to learn meaningful task-relevant features to some extent before introducing the TDP loss.
%
%
\section{Implementation details for previous works}
\label{sec:previous_details}
%
%

\paragraph{TDSR~\cite{sr_tdsr}.}
In our re-implementation, we use TDSR-0.01, which does not initially employ the task loss in the first one-third of the training process and then introduces the task loss with a ratio of 0.01 in the remaining training iteration.
Unlike the original paper, which used confidence loss and localization loss as task loss, we adopt the cross-entropy loss as task loss, as our ablation studies cover semantic segmentation and image classification.

\paragraph{SOD-MTGAN~\cite{bai2018sod}.}
Following the original paper, we introduce an additional fully connected layer in the task-specific head module $H_{\boldsymbol{\theta}_{\text{head}}}$, and utilize the output from that branch as the discriminator output.
Similarly to the TDSR, we replace the detection losses used in the original paper with the cross-entropy loss to cover semantic segmentation and image classification.
In the original paper, the authors set the loss weights for adversarial loss, task loss, and pixel-wise reconstruction loss as 0.001, 0.01, and 1.0, respectively.
However, we found that this setting resulted in significantly lower performance in our experiments.
Hence, we adjusted the task loss weight from 0.01 to 1.0.

\section{More visualization comparison results}
\label{sec:more_visualization}
\begin{figure*}
    \centering
    \captionsetup[subfigure]{labelfont=scriptsize, textfont=scriptsize}
    \renewcommand{\wp}{0.325}
        \subfloat[$\mathbb{I}_\text{LR}~{\rightarrow}~T$]{\includegraphics[width=\wp\linewidth]{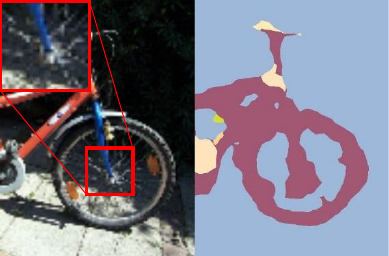}}
        \hspace{1mm}
        \subfloat[$S~{\rightarrow}~T$]{\includegraphics[width=\wp\linewidth]{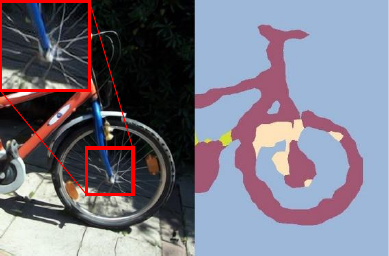}}
        \hspace{1mm}
        \subfloat[$T~{\rightarrow}~S$]{\includegraphics[width=\wp\linewidth]{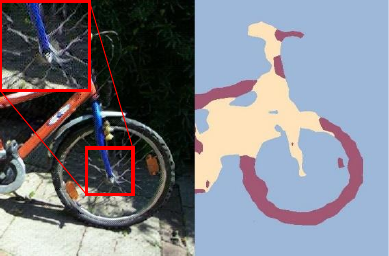}}
    \\
        \subfloat[$S~+~T$]{\includegraphics[width=\wp\linewidth]{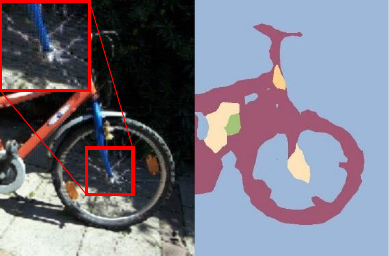}}
        \hspace{1mm}
        \subfloat[\textbf{SR4IR (Ours)}]{\includegraphics[width=\wp\linewidth]{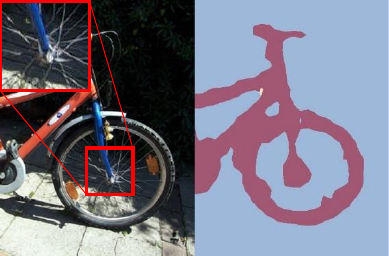}}
        \hspace{1mm}
        \subfloat[Ground-truth]{\includegraphics[width=\wp\linewidth]{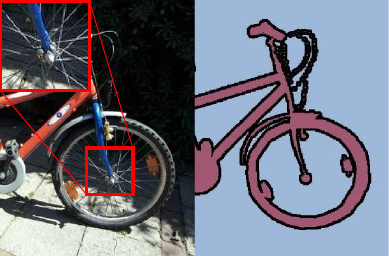}}
    \addtocounter{subfigure}{-6}
    \\
    \vspace{1cm}
        \subfloat[$\mathbb{I}_\text{LR}~{\rightarrow}~T$]{\includegraphics[width=\wp\linewidth]{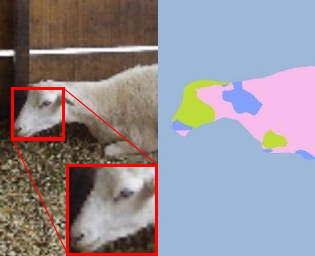}}
        \hspace{1mm}
        \subfloat[$S~{\rightarrow}~T$]{\includegraphics[width=\wp\linewidth]{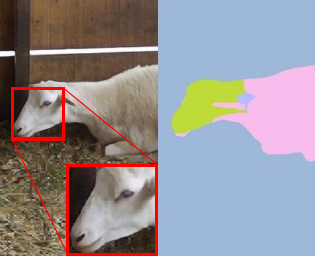}}
        \hspace{1mm}
        \subfloat[$T~{\rightarrow}~S$]{\includegraphics[width=\wp\linewidth]{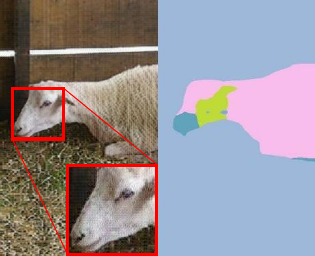}}
    \\
        \subfloat[$S~+~T$]{\includegraphics[width=\wp\linewidth]{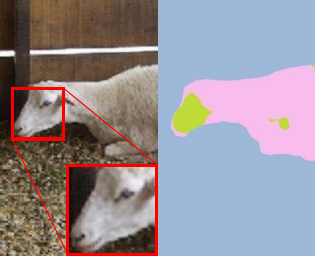}}
        \hspace{1mm}
        \subfloat[\textbf{SR4IR (Ours)}]{\includegraphics[width=\wp\linewidth]{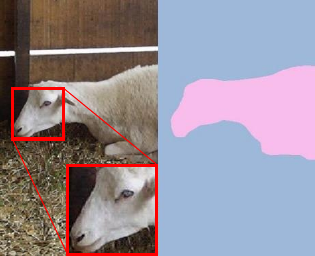}}
        \hspace{1mm}
        \subfloat[Ground-truth]{\includegraphics[width=\wp\linewidth]{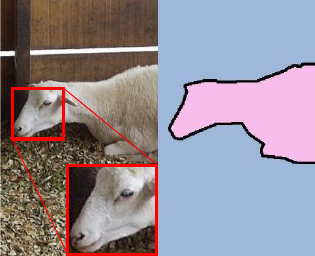}}
    \addtocounter{subfigure}{-6}
    \\
    \caption{\textbf{Visualization of images and semantic segmentation results on PASCAL VOC dataset~\cite{pascal-voc-2012}.}
        We present the restored images and the corresponding predicted segmentation maps.
        For (b), (c), (d), and (e), we use the SwinIR~\cite{sr_swinir} model with an SR scale factor of x4.
    }
    \vspace{1cm}
    \label{fig:more_qual_seg}
\end{figure*}

\begin{figure*}
    \centering
    \captionsetup[subfigure]{labelfont=scriptsize, textfont=scriptsize}
    \renewcommand{\wp}{0.32}
        \subfloat[$\mathbb{I}_\text{LR}~{\rightarrow}~T$]{\includegraphics[width=\wp\linewidth]{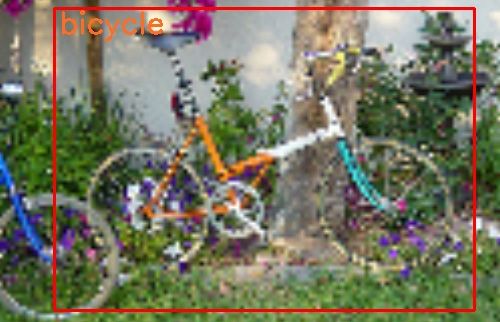}}
        \hspace{1mm}
        \subfloat[$S~{\rightarrow}~T$]{\includegraphics[width=\wp\linewidth]{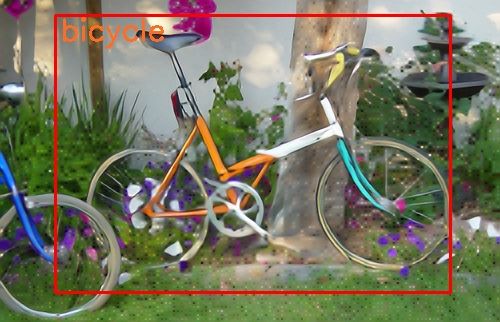}}
        \hspace{1mm}
        \subfloat[$T~{\rightarrow}~S$]{\includegraphics[width=\wp\linewidth]{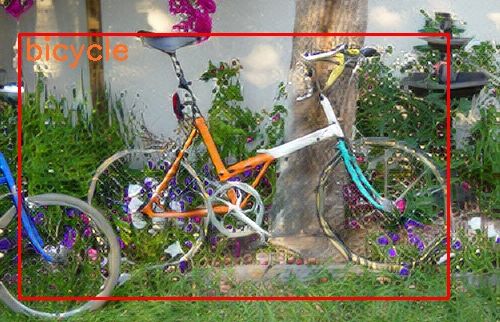}}
    \\
        \subfloat[$S~+~T$]{\includegraphics[width=\wp\linewidth]{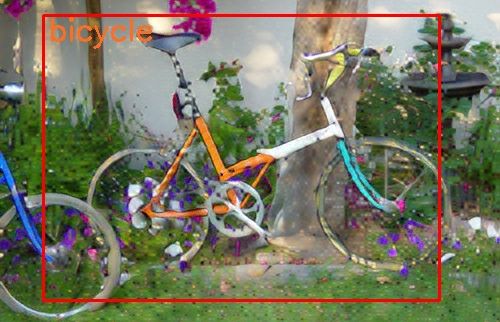}}
        \hspace{1mm}
        \subfloat[\textbf{SR4IR (Ours)}]{\includegraphics[width=\wp\linewidth]{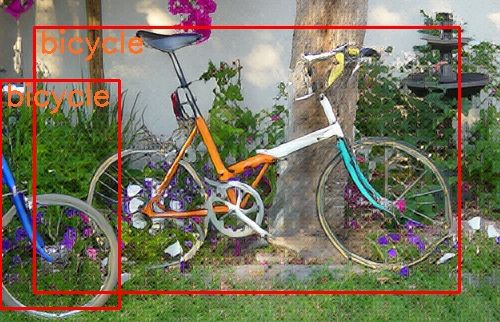}}
        \hspace{1mm}
        \subfloat[Ground-truth]{\includegraphics[width=\wp\linewidth]{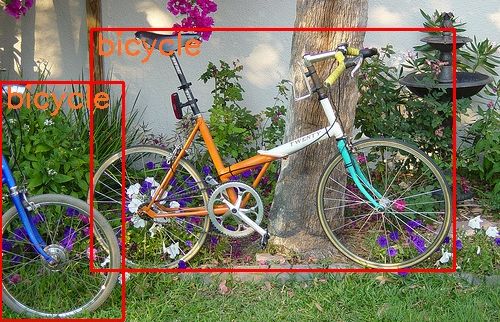}}
    \addtocounter{subfigure}{-6}
    \\
    \vspace{5mm}
    \renewcommand{\wp}{0.32}
        \subfloat[$\mathbb{I}_\text{LR}~{\rightarrow}~T$]{\includegraphics[width=\wp\linewidth]{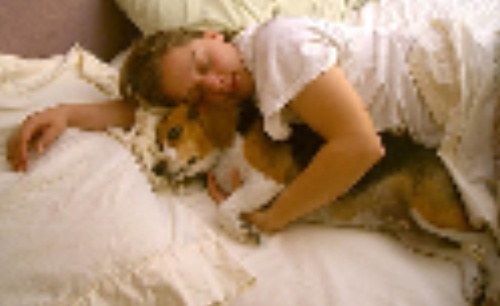}}
        \hspace{1mm}
        \subfloat[$S~{\rightarrow}~T$]{\includegraphics[width=\wp\linewidth]{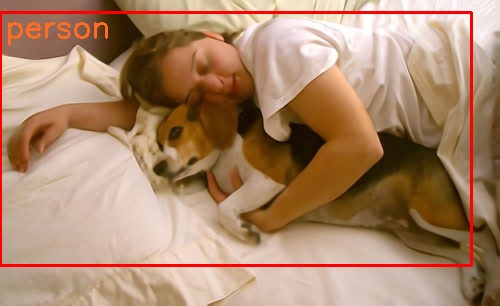}}
        \hspace{1mm}
        \subfloat[$T~{\rightarrow}~S$]{\includegraphics[width=\wp\linewidth]{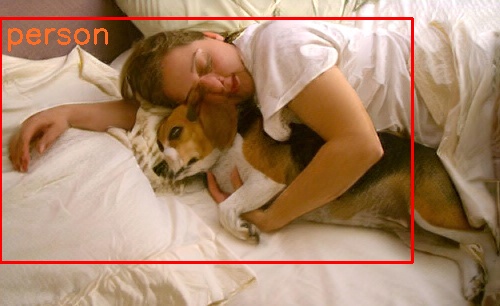}}
    \\
        \subfloat[$S~+~T$]{\includegraphics[width=\wp\linewidth]{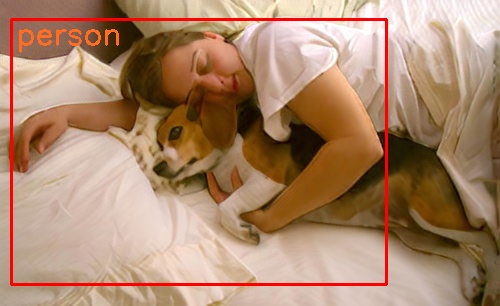}}
        \hspace{1mm}
        \subfloat[\textbf{SR4IR (Ours)}]{\includegraphics[width=\wp\linewidth]{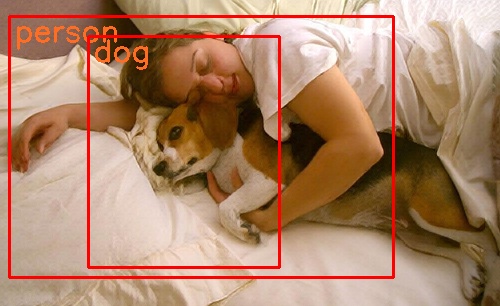}}
        \hspace{1mm}
        \subfloat[Ground-truth]{\includegraphics[width=\wp\linewidth]{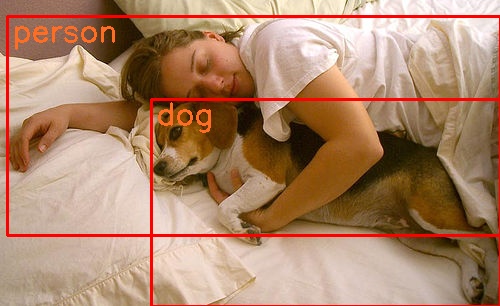}}
    \addtocounter{subfigure}{-6}    
    \\
    \vspace{5mm}
    \renewcommand{\wp}{0.32}
        \subfloat[$\mathbb{I}_\text{LR}~{\rightarrow}~T$]{\includegraphics[width=\wp\linewidth]{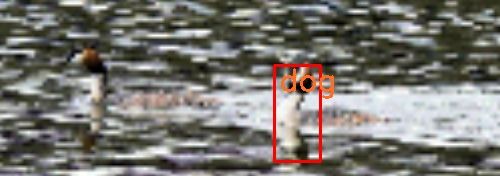}}
        \hspace{1mm}
        \subfloat[$S~{\rightarrow}~T$]{\includegraphics[width=\wp\linewidth]{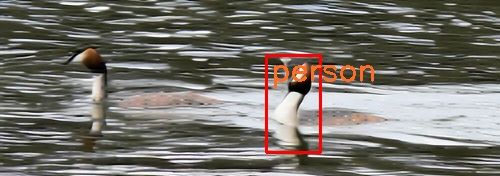}}
        \hspace{1mm}
        \subfloat[$T~{\rightarrow}~S$]{\includegraphics[width=\wp\linewidth]{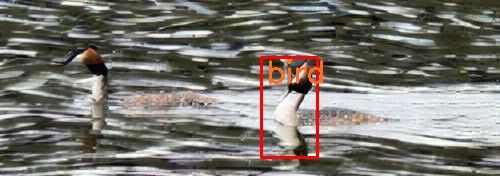}}
    \\
        \subfloat[$S~+~T$]{\includegraphics[width=\wp\linewidth]{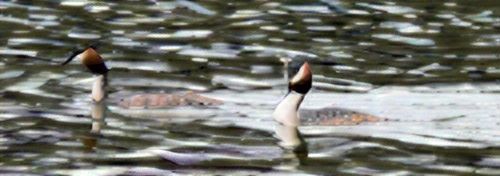}}
        \hspace{1mm}
        \subfloat[\textbf{SR4IR (Ours)}]{\includegraphics[width=\wp\linewidth]{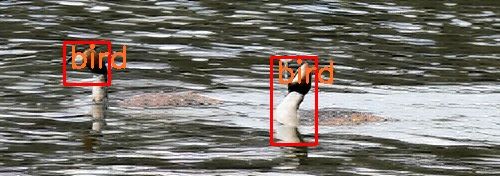}}
        \hspace{1mm}
        \subfloat[Ground-truth]{\includegraphics[width=\wp\linewidth]{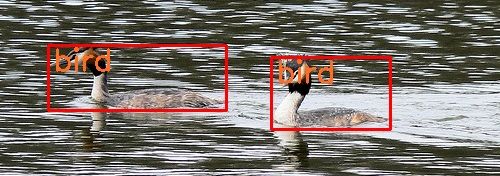}}
    \addtocounter{subfigure}{-6}
    \\
    \vspace{-0.2cm}
    \caption{\textbf{Visualization of object detection results on PASCAL VOC dataset.}
        The red box with orange annotation means the predicted object bounding box with the corresponding prediction.
        For (b), (c), (d), and (e), we use the SwinIR model with an SR scale factor of x4.
    }
    \vspace{-0.4cm}
    \label{fig:more_qual_det}
\end{figure*}

\begin{figure*}
    \centering
    \captionsetup[subfigure]{labelfont=scriptsize, textfont=scriptsize}
    \renewcommand{\wp}{0.32}
        \subfloat[$\mathbb{I}_\text{LR}~{\rightarrow}~T$]{\includegraphics[width=\wp\linewidth]{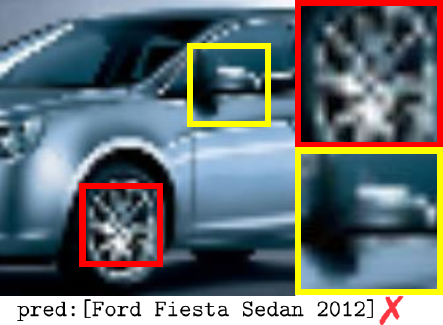}}
        \hspace{1mm}
        \subfloat[$S~{\rightarrow}~T$]{\includegraphics[width=\wp\linewidth]{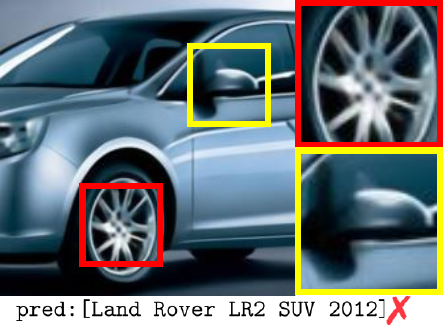}}
        \hspace{1mm}
        \subfloat[$T~{\rightarrow}~S$]{\includegraphics[width=\wp\linewidth]{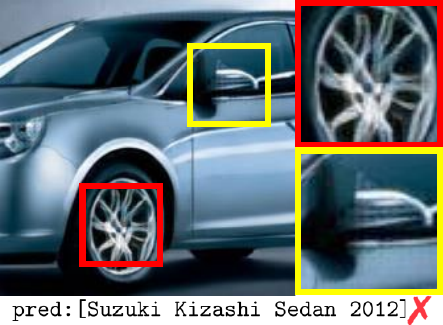}}
    \\
        \subfloat[$S~+~T$]{\includegraphics[width=\wp\linewidth]{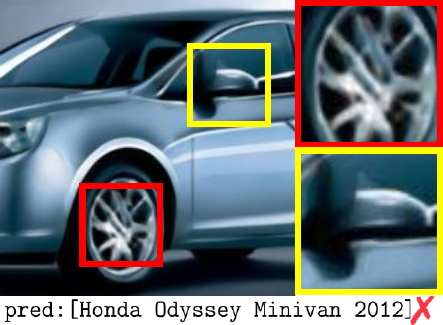}}
        \hspace{1mm}
        \subfloat[\textbf{SR4IR (Ours)}]{\includegraphics[width=\wp\linewidth]{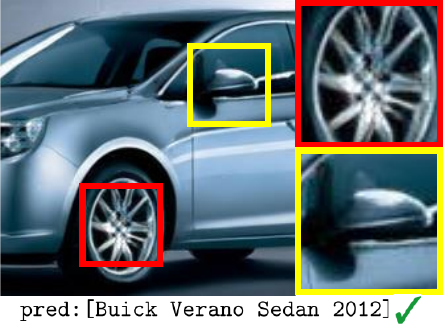}}
        \hspace{1mm}
        \subfloat[Ground-truth]{\includegraphics[width=\wp\linewidth]{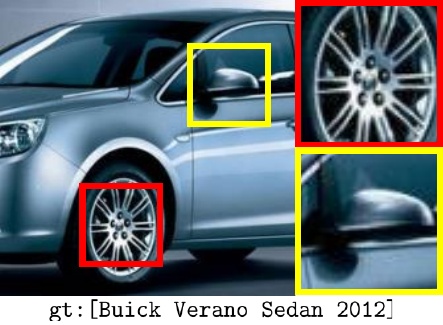}}
    \\
    \vspace{-0.2cm}
    \caption{\textbf{\textbf{Visualization of images and image classification results on StanfordCars~\cite{krause20133d} dataset.}}
        We present the restored images and the corresponding caption.
        The caption below the image represents the predicted image classification results, and a checkmark indicates if the prediction is correct.
        For (b), (c), (d), and (e), we use the SwinIR model with an SR scale factor of x4.
    }
    \label{fig:more_qual_cls_sc}
\end{figure*}

\begin{figure*}
    \centering
    \captionsetup[subfigure]{labelfont=scriptsize, textfont=scriptsize}
    \renewcommand{\wp}{0.32}
        \subfloat[$\mathbb{I}_\text{LR}~{\rightarrow}~T$]{\includegraphics[width=\wp\linewidth]{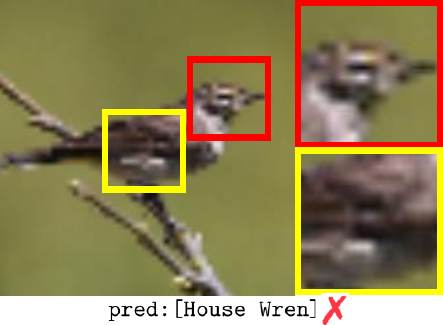}}
        \hspace{1mm}
        \subfloat[$S~{\rightarrow}~T$]{\includegraphics[width=\wp\linewidth]{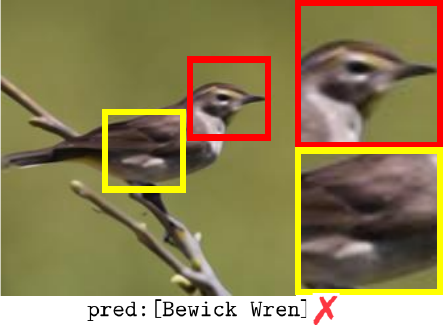}}
        \hspace{1mm}
        \subfloat[$T~{\rightarrow}~S$]{\includegraphics[width=\wp\linewidth]{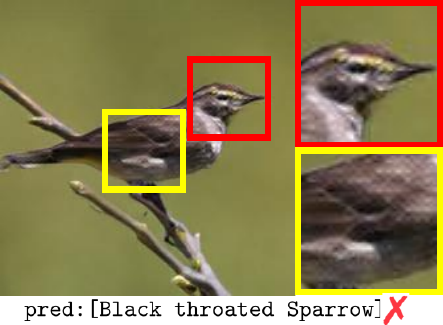}}
    \\
        \subfloat[$S~+~T$]{\includegraphics[width=\wp\linewidth]{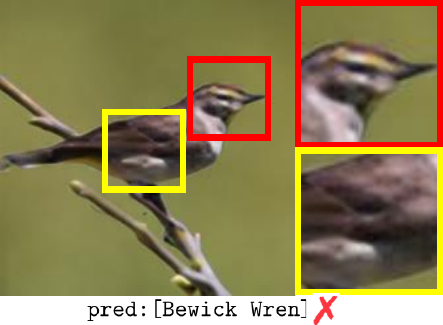}}
        \hspace{1mm}
        \subfloat[\textbf{SR4IR (Ours)}]{\includegraphics[width=\wp\linewidth]{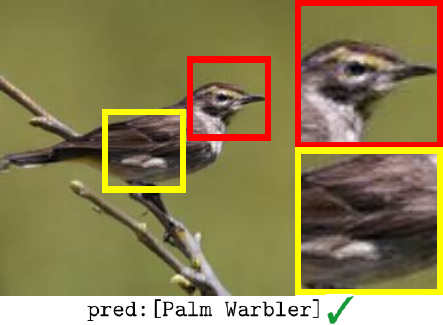}}
        \hspace{1mm}
        \subfloat[Ground-truth]{\includegraphics[width=\wp\linewidth]{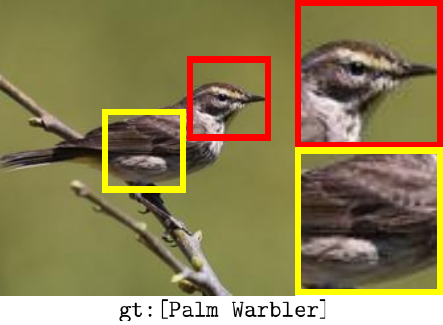}}
    \\
    \vspace{-0.2cm}
    \caption{\textbf{\textbf{Visualization of images and image classification results on CUB-200-2011 dataset~\cite{WahCUB_200_2011}.}}
        We present the restored images and the corresponding caption.
        The caption below the image represents the predicted image classification results, and a checkmark indicates if the prediction is correct.
        For (b), (c), (d), and (e), we use the SwinIR model with an SR scale factor of x4.
    }
    \label{fig:more_qual_cls_cub}
\end{figure*}

%
We present additional qualitative results for semantic segmentation (Figure~\ref{fig:more_qual_seg}), object detection (Figure~\ref{fig:more_qual_det}), and image classification (Figure~\ref{fig:more_qual_cls_sc},~\ref{fig:more_qual_cls_cub}).
We compare our SR4IR with all baselines, $\mathbb{I}_\text{LR}\rightarrow{T}$, $S\rightarrow{T}$, $T\rightarrow{S}$, and $S~+~{T}$, as introduced in Section~\ref{ssec:eval} of our main manuscript.
These results demonstrate that our SR4IR framework achieves the most accurate predictions across all tasks while producing visually pleasing results.
\begin{table}[!t]
\footnotesize
\centering
\setlength\tabcolsep{1.0pt}
\def\arraystretch{1.1}
\resizebox{1.0\linewidth}{!}{
    \begin{tabular}{L{4.0cm}|C{1.6cm}C{1.6cm}C{1.6cm}C{1.6cm}}
    \toprule
    \, & Set5 & Set14 & B100 & Urban100 \\
    \hline
    \;~EDSR-baseline & 32.10~/~0.863 & 28.58~/~0.743 & 27.56~/~0.711 & 26.04~/~0.763 \\
    \;~EDSR-baseline-SR4IR$_\text{Segmentation}$ & 29.32~/~0.800 & 26.76~/~0.688 & 26.39~/~0.660 & 23.77~/~0.667 \\
    \bottomrule
    \end{tabular}
}
\vspace{-0.2cm}
\caption{
    \textbf{PSNR~/~SSIM on x4 SR benchmark datasets.}
    For the task-driven SR network, we use EDSR-baseline~\cite{sr_edsr} trained by our SR4IR on the segmentation task.
}
\vspace{-0.45cm}
\label{table:benchmark}
\end{table}

%

\end{document}